\documentclass[letterpaper]{article} 
\usepackage{aaai24}  
\usepackage{times}  
\usepackage{helvet}  
\usepackage{courier}  
\usepackage[hyphens]{url}  
\usepackage{graphicx} 
\usepackage{subcaption}

\usepackage{paralist}
\usepackage[utf8]{inputenc} 
\usepackage[T1]{fontenc}    
\usepackage{url}            
\usepackage{booktabs}       
\usepackage{amsfonts}       
\usepackage{nicefrac}       
\usepackage{microtype}      
\usepackage{paralist} 
\usepackage[english]{babel}

\usepackage{csquotes}

\usepackage{amsmath}
\usepackage{amssymb}
\usepackage{placeins}
\usepackage{subcaption} 
\usepackage{mwe}
\usepackage{caption}
\usepackage{subcaption}
\usepackage[dvipsnames]{xcolor}
\usepackage{tikz,tkz-euclide,pgfplots}
\pgfplotsset{compat=1.15}

\usepackage{natbib}  
\usepackage{caption} 
\usepackage{algorithm}
\usepackage{algorithmic}

\usepackage{newfloat}
\usepackage{listings}
\DeclareCaptionStyle{ruled}{labelfont=normalfont,labelsep=colon,strut=off} 
\floatstyle{ruled}
\newfloat{listing}{tb}{lst}{}
\floatname{listing}{Listing}
\title{Learning Hybrid Dynamics Models with Simulator-Informed Latent States}
\author {
	Katharina Ensinger\textsuperscript{\rm{1,2}},
	Sebastian Ziesche\textsuperscript{\rm1},
	Sebastian Trimpe\textsuperscript{\rm2}
}
\affiliations {
	\textsuperscript{\rm 1} Bosch Center for Artificial Intelligence, Renningen, Germany \\
	\textsuperscript{\rm 2} Institute for Data Science in Mechanical Engineering, RWTH Aachen Univeristy\\
	katharina.ensinger@bosch.com
}

\usepackage{bibentry}

\begin{document}

\maketitle

\begin{abstract}
	Dynamics model learning deals with the task of inferring unknown dynamics from measurement data and predicting the future behavior of the system.
	A typical approach to address this problem is to train recurrent models. 
	However, predictions with these models are often not physically meaningful.
	Further, they suffer from deteriorated behavior over time due to accumulating errors. 
	Often, simulators building on first principles are available being physically meaningful by design.
	However, modeling simplifications typically cause inaccuracies in these models.
	Consequently, hybrid modeling is an emerging trend that aims to combine the best of both worlds.  
	In this paper, we propose a new approach to hybrid modeling, where we inform the latent states of a learned model via a black-box simulator.
	This allows to control the predictions via the simulator preventing them from accumulating errors.
	This is especially challenging since, in contrast to previous approaches, access to the simulator's latent states is not available. 
	We tackle the task by leveraging observers, a well-known concept from control theory, inferring unknown latent states from observations and dynamics over time.  
	In our learning-based setting, we jointly learn the dynamics and an observer that infers the latent states via the simulator.
	Thus, the simulator constantly corrects the latent states, compensating for modeling mismatch caused by learning. 
	To maintain flexibility, we train an RNN-based residuum for the latent states that cannot be informed by the simulator. 
\end{abstract}


\section{Introduction} \label{section:intro}
Physical processes $\left(x_n\right)_{n=0}^N \in \mathbb{R}^{d_x}$ can often be described via a discrete-time dynamical system
\begin{equation}\label{eq:dyn}
\begin{aligned}
x_{n+1}= f(x_n).
\end{aligned}
\end{equation}
In practice, the dynamics $f$ are often unknown but measurements are available.  
Typically, it is not possible to measure the full state space but noisy measurements $\hat{y}_n$ for a function $g$ of the states can be obtained by sensors, thus
\begin{equation}\label{eq:obs}
\begin{aligned}
y_n & = g(x_n) \\
\hat{y}_n  &= y_n +\epsilon_n, \textrm{ with } \epsilon_n \sim \mathcal{N}(0,\sigma^2),
\end{aligned}
\end{equation}
and $g: \mathbb{R}^{d_x} \rightarrow  \mathbb{R}^{d_y}$.
Our general motivation here is to make accurate predictions for the future behavior of the measurable states $y_n$ in Eq.~\eqref{eq:obs}.
One common strategy to address this problem is to train recurrent architectures on $\hat{y}_n$ in Eq.~\eqref{eq:obs} \citep{HochSchm97, 10.5555/3305890.3305957}.
Analogous to the unknown system, they possess internal latent states.
Predictions are obtained by mapping the latent states to the measurable states $y_n$ (cf. Eq. \eqref{eq:obs}) via an observation model. 
While these methods are able to accurately reflect the system's transitions, they often lack physically meaningful predictions, e.g. by violating physical laws. 
Further, small model mismatches lead to accumulating errors over time in the latent states and thus, also in the predictions  \citep{DBLP:conf/iclr/ZhouLXHH018}.

To address these problems, it is often beneficial to include physical prior knowledge such as energy conservation, invariants etc. in the architecture \citep{chen2020symplectic, NEURIPS2019_26cd8eca}.  
For many systems, it is even possible to model a physics-based simulator for the system producing stand-alone predictions \citep{10.1145/3447814, 10.1145/3514228}.
By design, predictions from these models are physically meaningful.
Further, they typically do not suffer as much from error accumulation.
However, modeling simplifications or incomplete knowledge lead to inaccuracies in the model.
In order to combine the best of both worlds, hybrid modeling (HM) fuses physics-based simulations with learning-based approaches \citep{takeishi2021physicsintegrated, wehenkel2023robust, yin2021augmenting}.
Most hybrid modeling approaches rely on analytical descriptions of physics laws.
However, such simulators are often not available in practice. 
Instead, numerical simulators building on these laws are available that are incompatible with most HM approaches.
Further, simulator software is often proprietary.
As a consequence, the simulators are often black-box. 
More specifically, they provide approximations $\hat s \in \mathbb{R}^{d_s}$ to the true outputs $y$ (cf. Eq.~\eqref{eq:obs}). 
Any additional insight as access to the simulator's internal latent states or dynamics is not available.

We propose to extract the hidden information and dynamics of the simulator outputs. 
By informing the latent states of a learning-based model with this information, we prevent them from error accumulation and ensure that they are consistent with the simulator.
In contrast to our approach, previous approaches in the black-box HM case, typically fuse the final predictions to one common prediction. 
The simplest approach is fusing them additively via a residuum model \citep{Suhartono_2017},
while more sophisticated approaches e.g. leverage the spectral properties \citep{ensinger2023combining}.
Thus, the simulator has an influence on the final predictions but not on the evolution of the learning-based component. 
Instead, we directly control the root of accumulating errors in the predictions by correcting the latent states. 
This approach is also less restrictive than, e.g., a simple residuum model since it requires less informative simulator trajectories. 

On a technical level, we inform our latent states by leveraging observers. 
Observers are commonly used in control systems inferring unknown latent states from observations for a \emph{known} dynamics model over time \citep{observer, BERNARD2022224}. 
Here, we propose to leverage them in a HM scenario.
In particular, we learn the dynamics for latent states that can explain both, data and simulator via separate observation models.
These latent states are informed and controlled via the simulator by learning an observer that receives the simulator outputs as input. 
Intuitively, the observer constantly minimizes the error between predicted and measured simulator outputs.
Thus, it compensates for model mismatch and prevents the latent states from accumulating errors.
As a consequence, it also prevents the predictions from accumulating errors, since latent states are mapped to predictions via the observation model. 
With the observer architecture, we ensure physical behavior to some extent. 
This is due to the fact that we can only reconstruct latent states that are able to explain the physics-based simulator. 
 
To maintain the flexibility of recurrent architectures, we train an additional recurrent neural network (RNN)-based residuum to address parts of the system, where the simulator is not informative. 
The influence of both components can be balanced in the loss. 
Additionally, the learning-based residuum can be controlled, e.g. by forcing it to vanish over time modeling transient behavior. 
Thus, we obtain full control over all components of the model enabling the simulator to take over as much responsibility over the predictions as desired. 
In the experiments, we show that our model produces accurate short and long-term predictions and is on par or better than baselines.
In particular, it also outperforms a recurrent architecture that receives the simulator as control input \citep{SCHON202219, 10.1145/3447814}, especially when the simulator is only partially informative. 
We also derive the differences and advantages with respect to this architecture intuitively and mathematically and support the empirical findings.  
 
Our model has additional advantages. 
Due to its architecture, system and simulator dynamics are modeled simultaneously.
This information can be used to buffer missing simulator outputs.
Further, the method can be extended to the pure learning-based scenario. 
Here, we substitute the simulator with a robust but incomplete learning-based substitute.
As in the hybrid case, the substitute prevents the full system from accumulating errors. 
In summary, our contributions are:

\begin{compactitem}
	\item A new view on HM that aims to maximize the influence of a black-box simulator by splitting the latent dynamics into two parts. One is fully controlled by the simulator, the other can be regularized arbitrarily. 
    \item An observer-based architecture that informs and constantly corrects the latent states of a learning-based model via the simulator. Thus, preventing error accumulation.
    \item We achieve higher or equal accuracy than learning-based and hybrid baselines. 
\begin{figure*}[tb]
	\centering
	\includegraphics[width=0.92\textwidth]{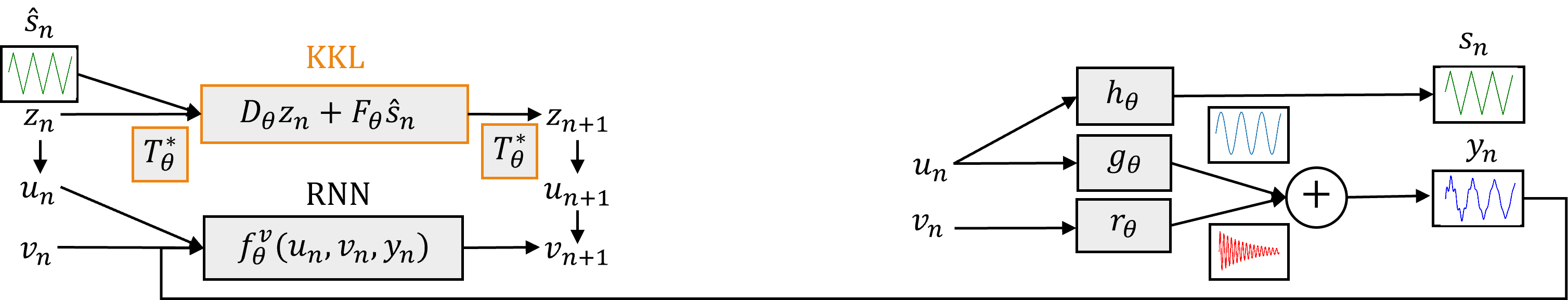}
	\caption{High-level overview of our method. Transition model (left): The simulator signal $\hat{s}$ is fed into a trainable observer to infer the OVS states $u$. 
	The non-OVS states $v$ are learned by an additional RNN. 
	Observation model (right): The simulator is reconstructed by $h_{\theta}$, while $g_{\theta}$ and $r_{\theta}$ reconstruct the measurements.}
	\label{fig:scheme}
\end{figure*} 
\end{compactitem}


\section{Background: Observer Design} \label{section:background}
In this work, we leverage observers, a concept from control theory in order to infer the unknown latent states of a learning-based model via a black-box simulator. 
Here, we provide the necessary background on observer design \citep{8472168,cdc-2019}.
Consider a system 
\begin{equation}\label{eq:simple}
\begin{aligned}
u_{n+1}&=f_u(u_n) \\
\hat s &=h(u_n),
\end{aligned}
\end{equation}
with dynamics $f_u:\mathbb{R}^{d_u} \rightarrow \mathbb{R}^{d_u}$, observation model $h:\mathbb{R}^{d_u} \rightarrow \mathbb{R}^{d_s}$ and measurements $\hat s$.
The latent states $u$ are denoted \emph{observable} in case they can be inferred from measurements  $\hat s$ over time, regardless of the initial condition, given known dynamics $f_u$.
The algorithm fulfilling this task is denoted observer. 
In contrast to the standard ML terminology, observable states are \emph{latent} and cannot be measured.

Partially observable systems denote systems, where only parts of the states can be reconstructed from the outputs via an observer \eqref{eq:observer} \citep{TAMI201654, Robenack2006}.  
 
\paragraph{Observer: }
The goal of an observer is to reconstruct the unknown but observable latent states $u$ from outputs $\hat s$ and known dynamics. Here, we refer to an observer as a mapping $\mathcal T$ that produces an estimate
$\tilde{u}$ with $\tilde{u}_{n+1}=\mathcal{T}(\hat s_1,\dots,\hat s_n,\tilde{u}_0)$ and
\begin{equation}\label{eq:observer}
|u_n-\tilde{u}_n| \rightarrow 0, n \rightarrow \infty. 
\end{equation}

\subsection{KKL observer}\label{sec:KKL}
In this work, we consider the so-called Kazantzis–Kravaris/Luenberger (KKL) observers due to their beneficial properties such as robustness against model mismatch \citep{DBLP:journals/corr/abs-2210-01476}. 
They can be designed in the continuous-time and discrete-time case \cite{8472168,cdc-2019}.
In the following, let $\mathcal U_0 \subset \mathcal U \subset  \mathbb{R}^{d_u}$  be a compact subset such that for
all initial conditions $u_0 \in \mathcal{U}_0$ and all $n \in \mathbb{N}$ it holds that $u_n \in \mathcal{U}$.
We will define the central properties of KKL observers here.
Mathematical details are added in Appendix Sec. 1.1. 
The key criterion for the existence of KKL observers is the so-called backward-distinguishability. 

\paragraph{Backward-distinguishability: }
Intuitively, for distinguishable trajectories in the latent space with $u_0^a \neq u_0^b$, there exists a point $t<0$ in the past, for which
the respective outputs are distinguishable as well, thus $\hat s_{t}^a \neq \hat s_{t}^b$.

\paragraph{Controllable pair: }
A matrix pair $(D,F)$ with $D \in \mathbb{R}^{d_u \times d_u}$ and $F \in \mathbb{R}^{d_u \times d_s}$ is denoted
controllable if the controllability matrix $C = [F, DF, D^2F,\dots,D^{d_u-1}F]$ has full row rank \citep{ogata1997modern}.

\paragraph{KKL observer: }
Define $d_z = d_y(d_u + 1)$.
Consider a backward-distinguishable system with dynamics $f_u$ and observation function $h$.
Furthermore, let $(D,F)$ be a controllable pair, where the eigenvalues $\lambda_1,\dots,\lambda_{d_z}$ of $D$ fulfill $\max(|\lambda_i|)<1$.
Then, under additional mild assumptions, there exists a continuous injective mapping
$T:\mathbb R^{d_u} \to \mathbb{R}^{d_z}$ with continuous pseudo-inverse
$T^{\star}:\mathbb R^{d_z} \to \mathbb{R}^{d_u}$ and
\begin{equation}\label{eq:KKL}
\begin{aligned}
T(f_u(u))=DT(u)+Fh(u)
\end{aligned}
\end{equation}
on $\mathcal U$ and 
\begin{equation} \label{eq:observation}
\lim_{n \rightarrow \infty} |u_n-T^{\star}(z_n)|=0
\end{equation}
for any trajectory defined via $z_{n+1}=Dz_n+F \hat s_n$.
Thus, $T^{\star}(z)$ is an observer for $u$.
 
\paragraph{Properties and limits: }
It is not clear how to obtain the transformation $T$ though its existence is guaranteed. 
\citet{9304163,9683485} approach the problem by sampling trajectories for $u$ and $z$ and solving regression problems for $T$. 
However, there is a significant difference since we consider the case of \emph{unknown} dynamics $f_u$.
Still, we benefit from the properties of KKL observers as we will demonstrate in the following sections.
We parametrize $T$ with a neural network (NN) similarly to \citet{9683277} that consider a learning-based but not a HM scenario. 

\section{Problem Setting} \label{section:problem}
In this section, we formulate our problem and demonstrate how the concepts in Sec. \ref{section:background} are related to it.
To this end, consider system \eqref{eq:dyn} with outputs $y_n$ (cf Eq.\eqref{eq:obs}).
Our goal is to make accurate predictions for $y_n$. 
Here, we consider the situation, where access to an inaccurate physics-based approximation $(\hat s_n)_{n=0}^N \in \mathbb{R}^{d_s}$ of the outputs is provided by a black-box simulator.
Thus, access to the simulator's latent states or the dynamics is not available. 
We propose to leverage the concepts in Sec. \ref{section:background} and learn how to reconstruct latent states from the simulator that are meaningful for the prediction task, thus maximizing the simulator's influence on the predictions. 
Since the simulator can typically not capture all dynamics, we introduce an additional residuum $r$. 

We model these concepts by splitting the latent state $x$ (cf Eq.~\eqref{eq:dyn}) into $u \in \mathbb{R}^{d_u}$ and $v \in \mathbb{R}^{d_v}$, where $d_x = d_u+d_v$.
With $u$, we denote the latent states that can be reconstructed from the simulator and refer to them as being \textbf{observable via the simulator (OVS)}.
The non-OVS states $v$ on the other hand cannot be reconstructed from the simulator.
We propose to formulate a common dynamics model for simulator and data by extending system \eqref{eq:partially_obs} via
\begin{equation}\label{eq:partially_obs}
\begin{aligned}
u_{n+1}&=f_u(u_n) \\
v_{n+1}&=f_v(u_n, v_n) \\ 
y_n & = g(u_n)+r(u_n,v_n) \\
\hat{s}_n & = h(u_n),
\end{aligned}
\end{equation}
with $f_u:\mathbb{R}^{d_u} \to \mathbb{R}^{d_u}$ and $f_v:\mathbb{R}^{d_u} \times \mathbb{R}^{d_v} \to \mathbb{R}^{d_v}$.
Here, $f_u$ determines the propagation of the OVS latent states $u$, while $f_v$ determines the propagation of the non-OVS latent states $v$. 
The observation model $g$ maps the OVS states $u$ to the OVS part of the predictions, while $r$ maps the OVS states $u$ and non-OVS states $v$ to the non-OVS residuum. 
The simulator is reconstructed via the observation model $h$.  
Our goal is to learn $f_u,f_v,g,h$ and $r$ from measurement data $\hat y$ and simulator outputs $\hat{s}$.

The key insight is that we force $u$ to be OVS by addressing $f_u$ via a trainable observer that receives $\hat s$ as an input.
By design, we obtain simulator-informed latent states $u$ and corresponding outputs $g(u)$.
In general, we can represent every system via Eq. \eqref{eq:partially_obs}.
By setting $d_v=0$ and removing the non-OVS residuum, we obtain the fully OVS case.  
By setting $d_u=0$, our model is reduced to the pure learning-based case. 
We choose an additive structure for the observation model in order to obtain control over the corresponding non-OVS residuum $r$.
However, different structures are also possible.

\section{Method: Hybrid KKL-RNN} \label{sec:method}
Here, we develop a hybrid model that extracts as much information as possible from black-box simulators and maximizes the simulator's influence on the predictions.
Thus, addressing the problem stated in Sec. \ref{section:problem}.
The key component is a trainable KKL observer that informs the OVS latent states of a learning-based model via the simulator.  
Since predictions are linked to latent states via an observation model, this allows to control the predictions via the simulator. 
In contrast to the standard setting in control, the dynamics in Eq. \eqref{eq:partially_obs} are unknown and have to be learned.
However, we can still benefit from the properties of KKL observers, in particular, we leverage their robustness against model errors here caused by learning \citep{DBLP:journals/corr/abs-2210-01476}. 
Fig. \ref{fig:scheme} provides an overview of our method. 
We model the evolution of the OVS states $u$ (cf. Eq. \eqref{eq:partially_obs}) by learning a KKL observer. Implicitly, this observer represents the dynamics $f_u$. 
This yields a state space that can explain the OVS parts of the data via the observation model $g$ and the simulator via the observation model $h$. 
Intuitively, the observer ensures that the estimated simulator outputs $h(u)$ coincide with the true simulator output $\hat s$ by constantly correcting the OVS states $u$. 
This prevents error accumulation in $u$ and thus, also in the OVS parts of the predictions $g(u)$. In order to obtain a flexible model, we address the non-OVS states by jointly training a residual model $f_v$ between measurements and the observer-based reconstructions.
We balance the influence of both components in the loss.

Fig. \ref{fig:scheme} also demonstrates the concept of OVS and non-OVS states.
Intuitively, the blue signal is OVS with simulator signal (green) since 
different points on the blue curve always correspond to different points on the green curve.
Thus, the blue states are backward-distinguishable via the green states and can be addressed by the trained KKL observer.
In contrast, identical outputs on the green curve yield different points on the red curve.
Due to the periodicity of the green states, the red states are not backward-distuingishable via the green states. 
Thus, the red signal is non-OVS and has to be addressed by the RNN residuum. 

\paragraph{Architecture: }
We build a flexible learning scheme based on the partially OVS system \eqref{eq:partially_obs}.
The OVS states $u \in \mathbb{R}^{d_u}$ are addressed by a trainable KKL observer.
To this end, we consider additional latent states $z$, that can be transformed into $u$ via $T^{\star}$ (cf. Sec. \ref{sec:KKL}). 
The non-OVS latent states $v \in \mathbb{R}^{d_v}$ are addressed
by RNN dynamics $f_{\theta}^v$.  
Consider training data $\hat{y}_n \in \mathbb{R}^{d_y}$ and simulator outputs $\hat{s}_n \in \mathbb{R}^{d_s}$. 
This yields
\begin{equation}\label{eq:transition}
\begin{aligned}
\begin{pmatrix}
z_{n+1} \\
u_{n+1}\\
v_{n+1}
\end{pmatrix}
=\begin{pmatrix}
D_{\theta}z_n+F_{\theta}\hat{s}_n \\
T^{\star}_{\theta}(z_n) \\
f_{\theta}^v(u_n,v_n,y_n)
\end{pmatrix}.
\end{aligned}
\end{equation}
and corresponding outputs
\begin{equation}\label{eq:observation}
\begin{pmatrix}
s_n  \\
y_n^v \\
y_n
\end{pmatrix}
= 
\begin{pmatrix}
h_{\theta}(u_n) \\
r_{\theta}(v_n)\\
g_{\theta}(u_n)+r_{\theta}(v_n)
\end{pmatrix}.
\end{equation}
Here, $\theta$ denotes the trainable parameters. 
We consider a trainable KKL observer
with matrices $D_{\theta} \in \mathbb{R}^{d_z \times d_z}, F_{\theta} \in \mathbb{R}^{d_z \times d_s}$ and a
mapping $T^{\star}_{\theta}:\mathbb{R}^{d_z} \rightarrow \mathbb{R}^{d_u}$.
Furthermore, we consider trainable observation models $h_{\theta}:\mathbb{R}^{d_u} \rightarrow \mathbb{R}^{d_s}$ $g_{\theta}^{u}:\mathbb{R}^{d_u} \rightarrow \mathbb{R}^{d_y}$
and $r_{\theta}:\mathbb{R}^{d_v} \rightarrow \mathbb{R}^{d_y}$.
Here, $h_{\theta}^{u}$ adresses the simulator, $g_{\theta}^{u}$ the OVS part of the predictions and $r_{\theta}$ the non-OVS residuum.
Due to the KKL structure, we learn the dynamics $f_u$ only implicitly by propagating $z$ through time and mapping to $u$ via $T^{\star}_{\theta}$. 
In case direct access to $f_u(u_n)=u_{n+1}=T_{\theta}^{*}(DT_{\theta}u_n+F \hat s_n)$ is required, we provide the option to jointly learn $T_{\theta}$.

In the experiments, we show that our method can also be leveraged in the pure learning-based scenario.
In particular, $\hat s$ is replaced by a robust but incomplete learning-based substitute.   
Similar to the HM scenario, the observer architecture allows the substitute to inform the latent states of a second high-accuracy model, preventing error accumulation. 
\paragraph{Models: }
To respect the requirements in Sec. \ref{section:background}, the matrix $D_{\theta}$ is modeled as a diagonal matrix with trainable bounded eigenvalues, while $F_{\theta}$ is chosen as a matrix with ones as entries.
This is not a restriction since a mapping $T$ exists for every controllable pair.
In the default scenario, $T_{\theta}^{\star}$ is modeled by an MLP similarly to \citet{9683277}.
However, our framework also offers the option to model $T_{\theta}^{\star}$ with an invertible NN consisting of affine coupling layers \citep{DBLP:conf/iclr/DinhSB17}.
The inversion provides access to $T_{\theta}$ and thus, to $f_u$.  
The trainable observation models
$g_{\theta}, h_{\theta}$ and $r_{\theta}$ are modeled as linear layers.
Here, we model the non-OVS part $f_{\theta}^v$ in Eq.~\eqref{eq:transition} with a gated recurrent unit (GRU) \citep{DBLP:conf/emnlp/ChoMGBBSB14}.
Details are provided in Appendix Sec. 1.2. 
\paragraph{Loss: }
Consider measurement data $\hat{y}_{0:N}$ and simulator outputs $\hat{s}_{0:N}$. 
The first $R$ steps $\hat{y}_{0:R}$ are used as a warmup phase for the non-OVS residuum to obtain appropriate
latent states. We train our model by computing an $N$-step rollout $z_{0:N}, y_{0:N}^v$ and $y_{0:N}$ via Eq.
\eqref{eq:observation} and minimizing the loss  
\begin{equation}
\hat{\theta}= \arg \min_{\theta} \Vert y_{0:N}-\hat{y}_{0:N} \Vert_2 + \Vert s_{0:N} - \hat{s}_{0:N} \Vert_2+\lambda \Vert y_{0:N}^v \Vert_2,  
\end{equation}
where $\Vert \cdot \Vert_2$ denotes the MSE. 
We introduce a regularization factor $\lambda \in \mathbb R$ that allows to balance the influence of the learning-based component as it is typical for hybrid models \citep{takeishi2021physicsintegrated, yin2021augmenting}.
This can yield some additional performance.
However, it is not mandatory and we do not include it in all experiments. 
In summary, the loss optimizes for a model that is able to represent the simulator outputs $\hat s$ via the
observation model $h_{\theta}$ and the measurements $\hat{y}$ via a residuum model $g_{\theta}(u)+r_{\theta}(u,v)$.

\paragraph{OVS and non-OVS components: }
Latent states that are OVS can be informed and corrected via the simulator over time. 
However, we do not intend to reconstruct the unknown internal states of the simulator as they might not even be informative enough for the data. 
Instead, OVS states can explain both, the simulator and parts of the data via different observation models.  
Also, there is no direct physical interpretation of the states.
However, since they are extracted by a physics-based simulator and respect backward-distinguishability, unphysical behavior is prohibited to some extent. 
We will also demonstrate that in the experiments. 
 
In order to maximize the influence of the simulator, we aim to maximize the influence of the OVS part $g$. 
The residual structure of the model allows to control the non-OVS counterpart $r_{\theta}$ (cf. Eq.~\eqref{eq:observation}).
As an example, it is easily possible to simulate transient behavior via a decaying residuum. This yields a model that is fully driven by the simulator after some time (cf. Eq.~\eqref{eq:transition}) via $g$. Thus, potential drifts or errors in the RNN do not affect the model performance in the long term.
In the experiments, we apply exponential damping by bounding the RNN observation model $r_{\theta}$ with an appropriate activation function and multiplying it with $\textrm{exp}(-a t)$. However, more elaborate strategies such as stable networks \citep{Schlaginhaufen2021LearningSD} could be easily incorporated.

%

\paragraph{Distinction to other architectures: }
In the experiments, we compare to GRUs that receive the simulator as additional control input \citep{SCHON202219, 10.1145/3447814}.
Under certain conditions, GRUs and other RNNs are a contraction \citep{BONASSI2021105049, miller2018stable}.
In these cases, they can act as an observer with latent states being driven by the simulator as desired.  
However, the GRU architecture is not restricted to these favorable models.  
Further, there is no guarantee that indeed a GRU with the required properties exists for the specific system.
For the KKL architecture, on the other hand, the existence is guaranteed if certain requirements described in Sec. \ref{section:background} hold.
By choosing the architecture described in Eq.~\eqref{eq:transition} and Eq.~\eqref{eq:observation}, the states $u$ are further OVS and the model is an observer by design under mild assumptions.
Additionally, the split in OVS and non-OVS part encourages the system to maximize the influence of the physics-based component.
The GRU architecture, on the other hand, could easily underestimate or even ignore the simulator, destroying the observer property.
Mathematical details for the statements are added in Appendix Sec. 1.2. 
In the experiments, we will show the advantage of the proposed architecture in practice. 
We will also demonstrate the advantages of addressing the OVS part in the partially OVS system ~\eqref{eq:partially_obs} indeed with an observer in contrast to modeling all components with separate GRUs.

Our model can also be interpreted as an extension of the standard residuum model by setting $h_{\theta}=g_{\theta}$.
We will show empirically that this extension allows to leverage simulator signals that are not informative enough for the standard residuum model and thus cause deteriorated behavior. 
Intuitively, this is due to the fact that we leverage hidden information from the simulator.

\section{Related Work} \label{section:rel}
HM is an emerging trend and a certain type of grey-box modeling. 
While general grey-box models often respect structural prior knowledge as energy-preservation, invariants etc. \citep{NEURIPS2019_26cd8eca, gaussprinciple2020geist, geist2021corl}, HM combines physics-based simulations with data-driven models.
Many works consider HM for dynamical systems or time-series.
Like us, \citet{Linial2020GenerativeOM} approach their HM task with RNNs.
In particular, they infer the initial states and the parameters of an ODE via an LSTM. 
However, in contrast to our setting, the system can be fully explained by the simulator with optimized parameters.
Another common approach in HM is extending a physics-based dynamics model, e.g. additively, with neural ODEs \citep{yin2021augmenting, qian2021integrating, 9337893} or modeling unknown parts of the dynamics with NNs \citep{SU1992327}.
\citet{rackauckas2021universal} present a unified view on these types of models, allowing to jointly learn physical and NN parameters. 
Recently, variational autoencoders are used to decode the latent states of observations and simulator from data and encode predictions from latent states and the physical model \citep{takeishi2021physicsintegrated}. 
Existing approaches are extended in \citep{wehenkel2023robust} via a data augmentation concept improving the behavior on unseen data. 
However, all these models assume direct access to the latent simulator states. 
In contrast, we consider black-box simulators that provide access only to output trajectories.

However, some approaches consider black-box simulators as well. 
One branch of HM with black-box simulators deals with optimizing parameters of the simulator \citep{DBLP:conf/iclr/RuizSC19, aushev2020likelihoodfree}.
However, this is not the setting that we consider since the simulator is fully informative once the parameters are adapted. 
A typical approach in our setting is to learn the errors or residua of simulator predictions and data \citep{Forssell97combiningsemi-physical, Suhartono_2017}.
Similar to our approach, \citet{ensinger2023combining} aim to control the long-term behavior of the predictions via the simulator.
To this end, they propose a complementary filtering approach. 
However, in contrast to their approach, our method is not restricted to simulators with correct low-frequency behavior. 
Furthermore, none of these works informs the latent states of the learning-based component. 
Another possibility is to provide the simulator as control input to an RNN \citep{SCHON202219, 10.1145/3447814}.
As discussed in detail in Sec. \ref{sec:method}, this can easily lead to an underestimation of the simulator. 

KKL observers have been combined with learning in different ways.  
In \citep{9304163}, nonlinear regression via NNs is performed in order to learn the nonlinear
transformation of a KKL observer. \citet{2204.00318} build up on the approach by optimizing the choice of
the controllable pair. \citet{9683485} propose a learning-based observer design by learning the nonlinear
transformation of a KKL observer with autoencoders. In contrast to our approach, all of these works consider
observer design for a dynamical system with \emph{known} dynamics.
However, some approaches consider KKL observers in the context of dynamics model learning.
\citet{buisson-fenet2023recognition} propose a KKL-based recognition model for neural ODEs. 
The initial latent state is obtained by running a KKL observer forward or backward. 
In contrast to our setting, the remaining rollout is not observer-based.
Furthermore, they do not consider HM. 
\citet{9683277} propose to construct an output predictor via a KKL observer.
The framework can be trained similar to standard recurrent architectures.
But in contrast to those architectures, mathematical guarantees for the output predictor can be obtained.
However, they do not consider HM. 
In contrast, we leverage the properties of KKL observers in order to control the behavior of the system by informing the latent states via the simulator.

\section{Experiments} \label{sec:exp}
\begin{figure*}[h!]
	\begin{minipage}{0.47\textwidth}
		\centering
		\includegraphics[width= \textwidth]{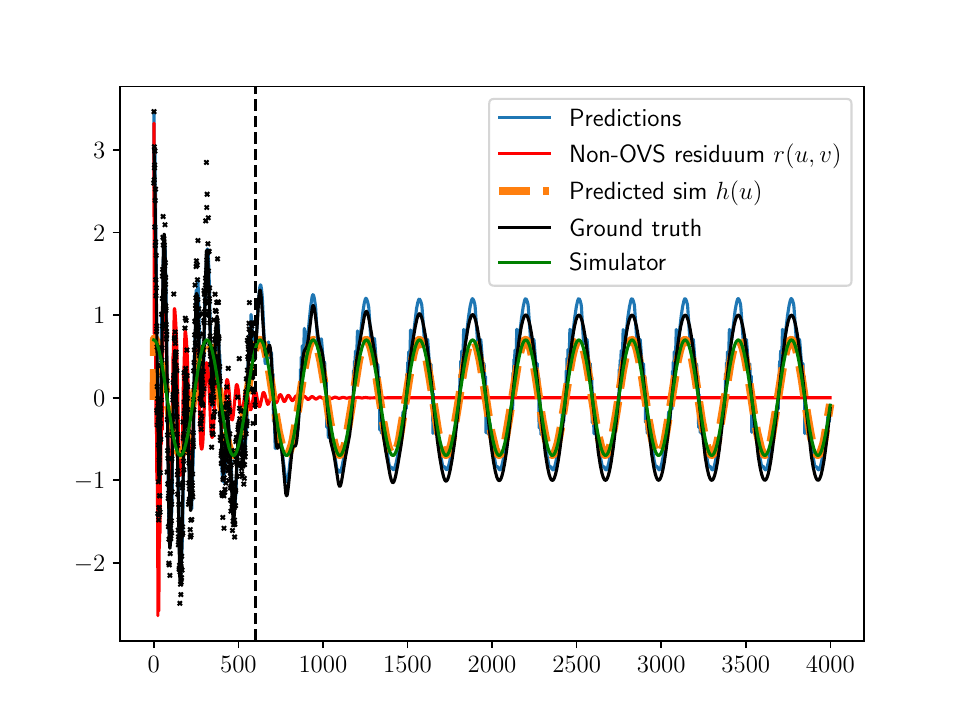}
	\end{minipage}	
	\hfill
	\begin{minipage}{0.47\textwidth}
		\centering
		\includegraphics[width= \textwidth]{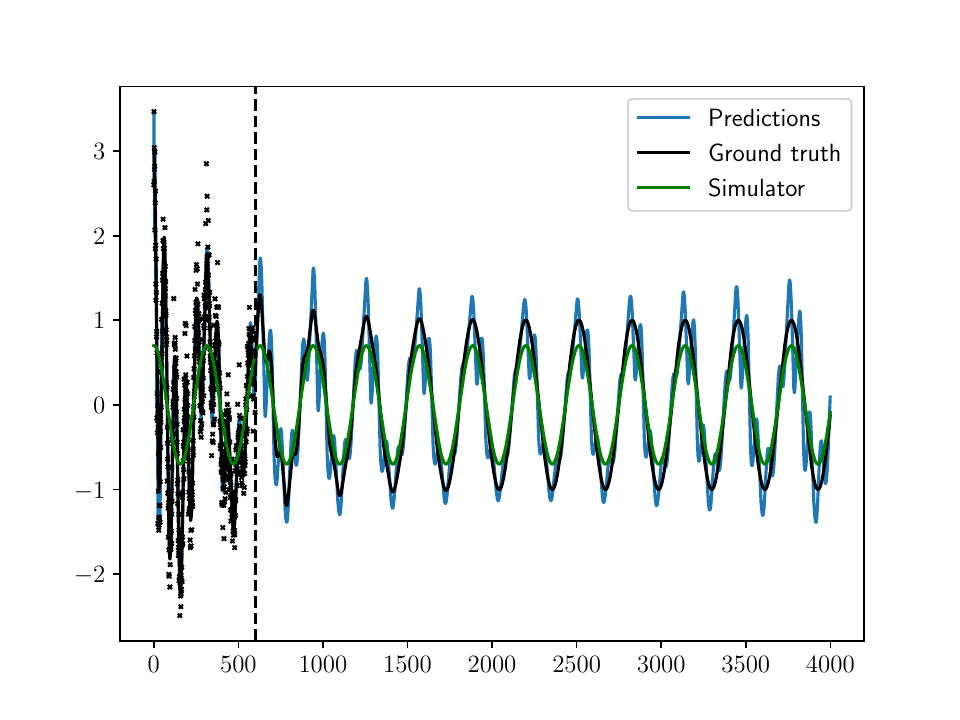}
	\end{minipage}	
	\caption{Rollouts over time for system i) with our hybrid KKL-RNN (left) and with the hybrid GRU (right). The training horizon is marked with dotted lines. The results demonstrate that our method reproduces all components correctly and learns  a plausible split in OVS and non-OVS. The hybrid GRU shows deteriorated and unphysical long-term behavior.}
	\label{fig:hybrid}
\end{figure*}

In this section, we show that: (i) Our KKL-RNN achieves equal or higher accuracy than baselines, especially in the partially OVS case; (ii) We learn a plausible split in OVS and non-OVS components with our method; (iii) Our method can buffer missing simulator inputs; (iv) We can easily incorporate properties as a decaying non-OVS part; (v) The concept can also be leveraged in the pure learning-based scenario.  

\paragraph{Baselines: } 
We model the non-OVS residuum in our KKL-RNN with GRU-based architectures.
Thus, our model can also be interpreted as a hybrid extension of a GRU.
In order to obtain baselines with comparable structure, we consider learning-based and hybrid baselines with a GRU backbone. 
We compare to the following baselines (for architecture details see Appendix Sec. 3.2).
\textbf{GRU: } State-of-the-art recurrent architecture for time-series and dynamics learning;  
\textbf{Hybrid GRU: } GRU with simulator trajectory as control input similar to \citep{10.1145/3447814};
\textbf{Residual model: } trains GRU predictions $r$ on the residual between data and simulator by minimizing $\Vert \hat y-(\hat s+r) \Vert_2$ similar to \citet{Forssell97combiningsemi-physical};
\textbf{Filter: } Fuses long-term information from the simulator with short-term information from a GRU by low-pass filtering the simulator and high-pass filtering the GRU \citep{ensinger2023combining}.
\textbf{Sim: } Physics-based simulator. 
For the pure learning-based task, we replace the simulator with a learning-based substitute.
\textbf{Ablation study: } In Appendix Sec. 2.2, we investigate how different aspects of the architecture affect the results. Thus, we consider several strategies to model the partially OVS system \eqref{eq:partially_obs} with GRUs. 

\paragraph{Learning task:}
For each experiment, we observe a single trajectory.
Rollouts are performed on a short part of the trajectory, while predictions are performed on the full trajectory. 
Either, we have access to real measurements or we consider simulated data corrupted with noise.
For the real-world data, we measure the root-mean-squared error (RMSE) between predictions $y$ and observations $\hat y$. For the simulated systems, we compare to the noise-free observations. 
All models are trained on batches of subtrajectories.
Here, we learn $T^{\star}_{\theta}$ with an MLP since direct access to $f_u$ is not required (cf. Sec. \ref{sec:method}).
See Appendix Sec. 3.2 for training details and Sec. 2.3 for invertible NN results.
 
\subsection{Systems}
For each of the four systems, we focus on different aspects of our method demonstrating that it can easily deal with different non-OVS residua (GRU, exponentially damped GRU) and missing simulator data. 
For equations of the simulated systems see Appendix Sec. 3.1.

\begin{figure*}[h!]
	\begin{minipage}{0.33\textwidth}
		\centering
		\includegraphics[width= \textwidth]{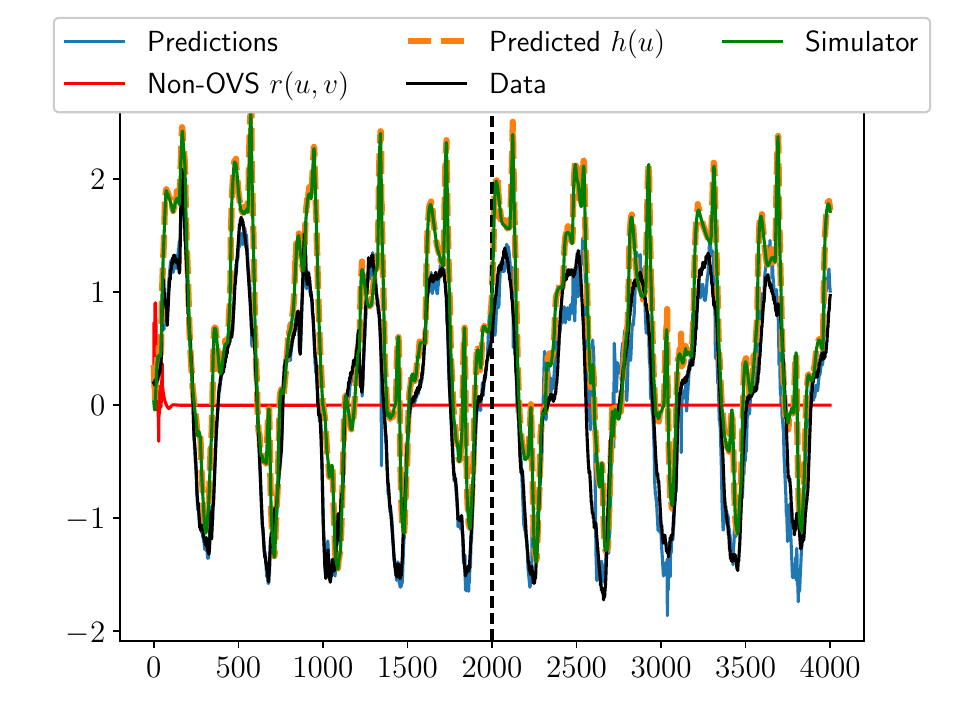}
	\end{minipage}	
	\hfill
	\begin{minipage}{0.33\textwidth}
		\centering
		\includegraphics[width= \textwidth]{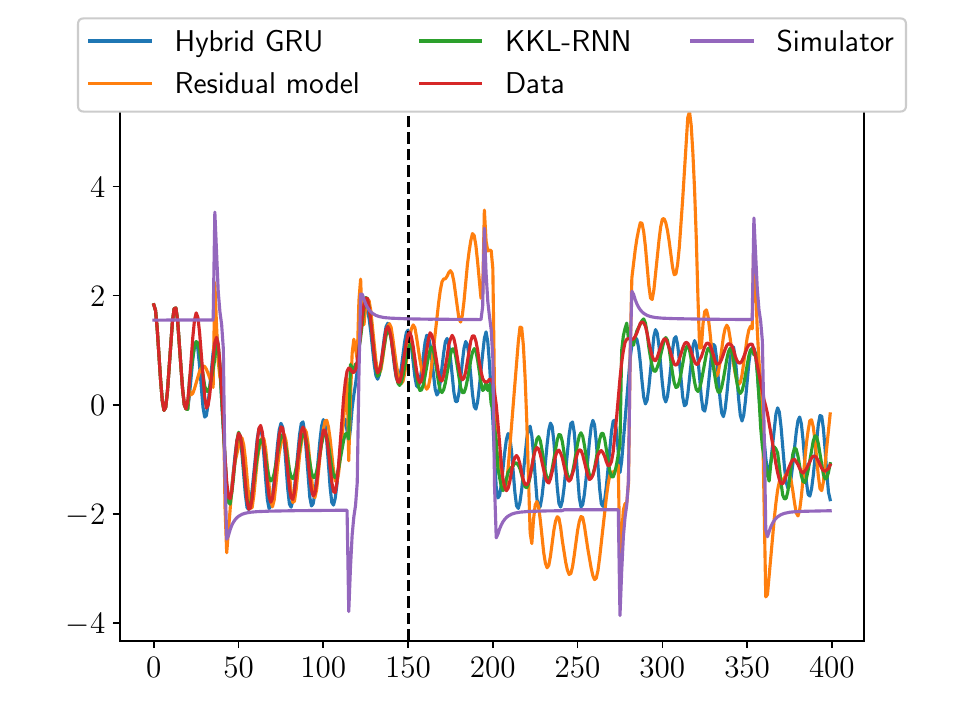}
	\end{minipage}	
    	\begin{minipage}{0.33\textwidth}
    	\centering
    	\includegraphics[width= \textwidth]{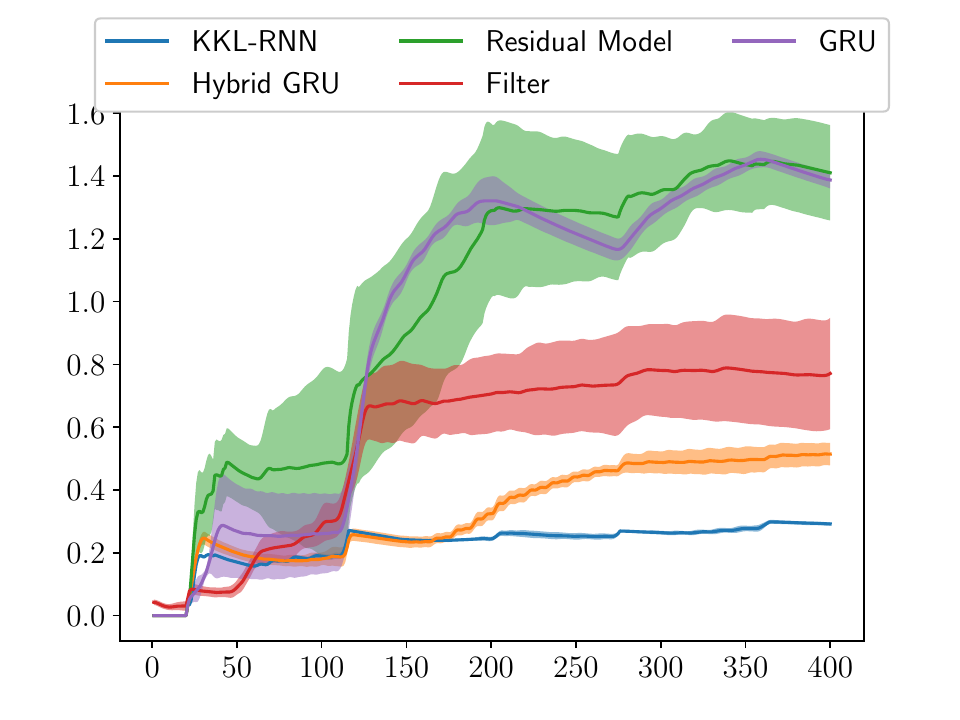}
    \end{minipage}	
	\caption{The rollouts with our KKL-RNN for System iii) (left) show that our method learns a plausible split into OVS and non-OVS and is able to buffer missing simulator inputs. The rollouts for System ii) (middle) demonstrate that our KKL-RNN produces more accurate results than the other HM approaches. The residual model even shows unphysical behavior.
	Accumulating errors in the baseline are further visible in the RMSE over time for System (ii) (right).}
	\label{fig:rest}
\end{figure*}	
\paragraph{i) Damped oscillations: }
Two superposed sine oscillations, where the second oscillation is damped and vanishes over time.
The simulator is represented by a sine wave with the correct main frequency but a modeling mismatch in the amplitude  (see Fig. \ref{fig:hybrid}). 
We enforce a vanishing residuum by exponentially damping the GRU observations $r_{\theta}$. 
Thus, after some time, the predictions are solely determined by the observer. 
The system is trained on a 600-steps interval, predictions are performed with 1500 steps. 

\paragraph{ii) Double-torsion pendulum: }
We consider the measurements and the corresponding numerical simulation from \citet{lisowski} and use the first 150 steps for training.
To add a transient component, we artificially add decaying sinusoidal oscillations to the data
(see. Fig. \ref{fig:rest} (right)).
We model the non-OVS residuum $r$ with a standard GRU.

\paragraph{iii) Drill-string system: }
We train on measurements provided in \citet{torsion} Fig. 14 and the corresponding simulator.
2000 time steps are used for training and 2000 additional time steps for predictions. 
To demonstrate the robustness of the method, we randomly remove 500 steps from the simulator during the prediction phase.  

\paragraph{iv) Van-der-Pol oscillator (pure learning-based): }
We extend our ideas and results to the pure learning-based scenario. 
In particular, the observer architecture is leveraged to fuse models with contrastive strengths.
Thus, we learn a simplified but robust model as a simulator substitute for the KKL-RNN.
As before, we model the non-OVS residua of our KKL-RNN with GRUs.
Here, we simulate a Van-der-Pol oscillator with external sine excitation.
As a simulator substitute, we consider a simple sine wave with parameters that are trained jointly with the models. 
Here, the Sim baseline is obtained by fitting the sine wave to the data. 

In Appendix Sec. 2.1, we consider a second example, that extends experiment ii) in \citet{ensinger2023combining} .
As in their setting, a GRU trained on a low-pass filtered signal serves as a simulator substitute.
In contrast to them we inform the high-pass components via the low-pass signal.


\paragraph{Results: }
\begin{table*}[!htpb]
	\centering 
	\begin{tabular}{rcccccc}
		\noalign{\smallskip}  \hline \noalign{\smallskip}
		task & GRU & Residual Model  & Hybrid GRU & Filter &  KKL-RNN (ours) & Sim \\ 
		\hline
		i) & 1.02 (0.03) & 0.31 (0.04) &  0.27 (0.08) &  0.64 (0.21) & \textbf{0.16} (0.03) & 0.36 \\
		ii)& 1.39 (0.03) & 1.41 (0.15) & 0.51 (0.04) &  0.61 (0.09)& \textbf{0.24} (0.02) & 1.09 \\
		iii) & 0.40 (0.19) & 0.63 (0.10) & \textbf{0.26} (0.01) & 0.60 (0.01) & \textbf{0.23} (0.01) & 0.66 \\
		iv) &  0.43 (0.19) & 0.51 (0.15) & 0.09 (0.065)&  - & \textbf{0.02} (0.001) & 0.537 \\
		\noalign{\smallskip} \hline \noalign{\smallskip}
	\end{tabular}\quad
	\caption{RMSEs for Systems i)-iii) (mean (std)) over 5 independent runs.}
	\label{t:hybrid}
\end{table*}

The results demonstrate that our method produces accurate long and short-term predictions.
Also, the simulator is reconstructed accurately (cf. Fig. \ref{fig:hybrid} and \ref{fig:rest} (left)).
This allows to buffer missing simulator information via the learned simulator signal as demonstrated for System iii). Further, the learned non-OVS residua $r$ indeed correspond to the non-OVS parts of the system, represented e.g. by transient behavior (cf. Fig. \ref{fig:hybrid} and \ref{fig:rest} (left)).
This property is leveraged for System i), where the hybrid GRU produces deteriorated and unphysical long-term behavior. 
Such behavior can not occur for the OVS components of our KKL-RNN since it violates backward-distinguishability. 
The non-OVS transient oscillations are further damped over time by design.
Table \ref{t:hybrid} shows that we achieve higher accuracy than the standard GRU, which suffers from deteriorated long-term behavior on all systems. 
Further, our method is also superior to the Residual model and Filter (cf. Fig. \ref{fig:rest} (middle)) since, in contrast to these approaches, the simulator informs the latent states of the learning-based component. 
This prevents unphysical behavior as it occurs in the Residual model (cf. Fig. \ref{fig:rest} (middle)) and allows to leverage less informative simulators.
As explained, informing the latent states further prevents error accumulation. 
The Filter also prevents error accumulation to some extent by adopting the long-term behavior of the simulator.
However, it relies on the assumption that the simulator provides the correct low-frequency information, which is not the case for the systems here (e.g. the simulator has the wrong amplitude in System (i)).
The accumulating errors in the baselines are further clearly visible in the accumulating RMSE over time (cf. Fig. \ref{fig:rest} (right)).
Often, the accuracy of our method is similar to the hybrid GRU (cf. System iii). 
It can be interpreted that in these cases, the GRU acts as an observer as explained in Sec. \ref{sec:method}.
However, for Systems i), ii) and iv), our method provides higher accuracy than the hybrid GRU.
A likely explanation is that, in contrast to our method, the hybrid GRU does not learn an optimal split into OVS and non-OVS components especially if the system is not fully OVS. 
This coincides with the findings that the GRU could ignore or underestimate the simulator input (cf. Sec. \ref{sec:method}).

System iv) shows that the findings extend to the pure learning-based scenario. 
Hybrid GRU and KKL-RNN outperform the other methods. 
Intuitively, both learn a sine wave with matching frequency. 
Further, they learn to infer the Van-der-Pol oscillator via it.
However, the KKL-RNN still outperforms the hybrid GRU that is not able to perfectly reproduce the oscillations. This indicates again that the hybrid GRU does not learn a fully-OVS system.
However, the results for the hybrid GRU can be leveraged.
They demonstrate that a standard GRU can be prevented from deteriorating long-term predictions by supporting it with a simple trainable signal.
Similar findings for another pure learning-based experiment are provided in Appendix Sec. 2.1. 

The ablation study in Appendix Sec. 2.2 demonstrates that the GRU architectures trained on Eq.~\eqref{eq:partially_obs} do not learn an optimal split into OVS and non-OVS and lack high accuracy.
In particular, the non-OVS residuum takes over parts that are actually OVS.
Since they are also trained on the partially OVS system \eqref{eq:partially_obs}, this suggests that the KKL observer is an essential component of our architecture. 
in Appendix Sec. 2.4, we provide additional plots, runtimes and plot the RMSE over time, indicating again accumulating errors in the baselines. 

\section{Conclusion}
We propose a hybrid modeling scheme that allows to extract hidden information from black-box simulators by informing the latent states of a learning-based model.
To this end, we train a KKL observer that infers OVS latent states via the simulator. 
The OVS states and corresponding predictions are thus constantly controlled by the simulator, preventing them from error accumulation.
Interesting aspects for future work are the extension to the stochastic setting or to different observers and partially observable forms.
\section*{Acknowledgements}
The authors thank Barbara Rakitsch and Mona Buisson-Fenet for valuable discussions.
\bibliography{bib}

\end{document}


\maketitle

\section{Mathematical background} \label{section:mathematics}
In this section, we will provide the necessary mathematical background and equations.
First, we will provide the background on KKL observers \citep{cdc-2019}.
Then, we will demonstrate in which cases the hybrid GRU could behave like an observer as described in the method and experimental section. 
We will also demonstrate, in which cases the properties are not fulfilled and the GRU is not able to act as an observer. 

\subsection{KKL observer} \label{section:background}
In this section, we add the necessary background on observer design.
In particular, we define backward distinguishability mathematically and list the full set of assumptions.
Analoguous to the background section, we consider a system with dynamics $f_u:\mathbb{R}^{d_u} \rightarrow \mathbb{R}^{d_u}$,
states $u$ and the observation model $h:\mathbb{R}^{d_u} \rightarrow \mathbb{R}^{d_s}$ with measurements $\hat s$.

\paragraph{Backward distinguishability: }
Consider a system with invertable dynamics $f_u$ and observation function $h$.  
Let $\mathcal{O}$ be an open bounded set containing $\mathcal{U}$.
Then, the system is denoted backward $\mathcal{O}$-distinguishable on $\mathcal{U}$ if
for any trajectories $u^a$ and $u^b$ with $(u^a_0, u^b_0) \in \mathcal{U}\times\mathcal{U}$ and $u^a_0 \neq u^b_0$,
there exists $T > 0$ such that $h(f_u^{-T}(u^a_0)) \neq h(f_u^{-T}(u^b_0))$ and
$(f_u^{-n}(u^b_0), f^{-n}(u^b_0)) \in \mathcal{O}\times\mathcal{O}$ for $n=0, \dots, T$.
A system is called backward distinguishable if there is such an $\mathcal O$.

\paragraph{Assumption 1: }
The dynamics function $f_u$ is invertible and $f_u^{-1}$ and $h$ are of class $C^1$ and globally Lipschitz.  
\paragraph{Assumption 2: }
The system with dynamics $f_u$ and observation function $h$ is backward-distinguishable.

Based on the assumptions the following theorem holds (cf. \citep{cdc-2019}).
\begin{theorem}[KKL-observer]\label{Thm1}
	Suppose assumptions 1 and 2 hold.
	Define $d_z = d_y(d_u + 1)$. 
	Let $c=\sup\{|(f^{-1})^{\prime}(u)| \mid u \in \mathcal{U}\}$ and $\mathcal{D}$ be the open disc in $\mathbb C$ of
	radius $\min\{1,1/c\}$. Then, there exists a set $S$ of zero measure in $\mathbb{C}^{d_z}$ such that for any
	diagonalizable matrix $D \in \mathbb R^{d_z \times d_z}$ with eigenvalues $(\lambda_1,\dots,\lambda_{d_z})$ in
	$\mathcal{D}^{d_z}\setminus S$ and any $F \in \mathbb{R}^{d_z \times d_s}$ such that $(D,F)$ is a controllable pair and 
	$\max(|\lambda_i|)<1,$
	there exists a continuous injective mapping $T:\mathbb R^{d_u} \rightarrow \mathbb{R}^{d_z}$ that satisfies the
	following equation on $\mathcal U$
	\begin{equation}
	\begin{aligned}
	T(f_u(u))=DT(u)+Fh(u),
	\end{aligned}
	\end{equation}
	and its continuous pseudo-inverse $T^{\star}:\mathbb C^{d_z} \rightarrow \mathbb{R}^{d_u}$ such that any trajectory of $z_{n+1}=Dz_n+F \hat s_n$ satisfies
	\begin{equation} \label{eq:observation}
	\lim_{n \rightarrow \infty} |u_n-T^{\star}(z_n)|=0.
	\end{equation}
	Thus, $T^{\star}(z)$ is an observer for $u$. 
\end{theorem}

\subsection{GRU architecture}
In this section, we will present the central results for GRUs that serve as a backbone and a baseline for our architecture. 
We will closely follow the notations and results in \citet{BONASSI2021105049}.
We will first recap the GRU dynamics and the warmup phase. 
The transition function $x_{n+1}=f(x_n, \tilde y_n, i_n)$ of a GRU is given as 
\begin{equation}\label{eq:GRUequation} 
\begin{aligned}
x_{n+1} &=z_n \circ x_n+(1-z_n) \circ \phi(\tilde W_r i_n+ W_r \tilde y_n+U_r h_n \circ x_n+b_r) \\
z_n &=\sigma(\tilde W_z i_n+W_z \tilde y_n+U_zx_n+b_z) \\
h_n& = \sigma(\tilde W_f i_n+W_f \tilde y_n+U_fx_n+b_f), 
\end{aligned}
\end{equation}
where $x_n \in \mathbb{R}^{d_x}$ is the state vector and $i_n \in \mathbb{R}^{d_s}$ is the control input.
In our case, the control input is given by the simulator $\hat s$. 
For the reconstructed observations $y \in \mathbb{R}^{d_y}$ it holds that 
\begin{equation}\label{eq:obs}
y_n=g(x_n)=U_ox_n
\end{equation}
with trainable matrix $U_0$. 
In the following, we specify the input $\tilde y_n \in \mathbb{R}^{d_y}$.  

\paragraph{GRU warmup phase}
A GRU ~\eqref{eq:GRUequation} is trained by feeding the measurements $\hat y_n \in \mathbb{R}^{d_y}$ as input on a small warmup phase of length $R$. 
After the warmup phase, the reconstructed observations are provided as inputs. 
This yields
\begin{equation}\label{eq:warmup}
\tilde y_n = 
\begin{cases}
\hat y_n & \textrm{ for } n \leq R, \\
y_n=U_o x_n & \textrm{ for } n > R. 
\end{cases}
\end{equation}

We will first demonstrate, under which conditions the system we consider could act as an observer.
Further, we will demonstrate conditions, under which it can not act as an observer. 
For both cases, we will derive some special cases. 
We will show, how a GRU could reproduce a partially OVS system and how it can reproduce a fully OVS system.
We will further demonstrate, how the GRU could ignore the simulator. 

In the following, we will focus on transitions after the warmup phase, thus $y_n = U_o x_n$.
Analogous to \citet{BONASSI2021105049} consider the following assumption.

\paragraph{Assumption 1: }
The initial state of the GRU network \eqref{eq:GRUequation} belongs to an arbitrary large but bounded set 
$\check{\mathcal X} \supseteq \mathcal X$, defined as 
\begin{equation}
\check{\mathcal X} = \{x \in \mathbb R^{d_n}: \Vert x \Vert_{\infty} \leq \check \lambda\},
\end{equation} 
with $\lambda \geq 1$. 

The central property of an observer is the independence of the initial conditions.
For the GRU this is the case, if the system forms a contraction.
Thus, we will show under which conditions the GRU is a contraction.  
The following results are based on \citet{BONASSI2021105049} adapted to our setting. 
\begin{lemma}[GRU properties]\label{lem:GRUproperties}
	Consider the GRU \eqref{eq:GRUequation}, two different initial values $x_a$ and $x_b$ after the warmup phase and identical control inputs $i$. 
	If 
	\begin{equation}
	\begin{aligned}
    \check \sigma_z & + (1-\check \sigma_z)\left(\Vert U_r \Vert_{\infty} (\frac{1}{4} \check \lambda \Vert U_f \Vert_{\infty}+\check \sigma_f)+\Vert W_r \Vert_{\infty}\Vert U_o \Vert_{\infty} 
    +\frac{1}{4} \check \lambda \Vert U_r \Vert_{\infty}\Vert W_f \Vert_{\infty}\Vert U_o \Vert_{\infty}\right)\\
    & +\frac{1}{4}(\check \lambda + \check \phi_r)(\Vert U_z \Vert_{\infty}+\Vert W_z \Vert_{\infty}\Vert U_o \Vert_{\infty})<1
  \end{aligned}
  \end{equation}
	with 
	\begin{equation}
	\begin{aligned}
	\check{\sigma}_z &=\sigma(\Vert W_z \ \check \lambda U_z \ b_z \Vert_{\infty}) \\
	\check{\sigma}_f & = \sigma(\Vert W_f \ \check \lambda U_f \ b_f \Vert_{\infty}) \\
	\check{\phi}_r &= \phi(\Vert W_r \ \check \lambda U_r \ b_r) \Vert_{\infty}),
	\end{aligned}
	\end{equation}
	then it holds that 
	\begin{equation}
	\Vert f(x_a, y_a, i)-f(x_b, y_b, i) \Vert \leq  C \Vert x_a-x_b \Vert,
	\end{equation}	
	where $C \in (0,1).$ 
\end{lemma} 
\begin{proof}
	Consider
	\begin{equation}
	\Delta x^+ = f(x_a, i, y_a)-f(x_b, i, y_b)
	\end{equation}
	and 
	\begin{equation}
	\Delta x = x_a-x_b.
	\end{equation}
	The proof of Theorem 2 in \citet{BONASSI2021105049} yields for the jth component of $\Delta x^+$
	\begin{equation}
	\begin{aligned}
	|\Delta x_j^+| & \leq \alpha_{\Delta_x} \Vert \Delta x \Vert_{\infty} + \alpha_{\Delta_u} \Vert U_o \Delta_x \Vert_{\infty} \\
	& \leq \alpha_{\Delta_x} \Vert \Delta x \Vert_{\infty} + \alpha_{\Delta_u}  \Vert U_o \Vert_{\infty} \Vert \Delta x \Vert_{\infty}
	\end{aligned}
	\end{equation}
	with 
	\begin{equation}
	\begin{aligned}
	   \alpha_{\Delta_x} & = z_{aj}+\frac{1}{4}(\check \lambda + \check \phi_r)\Vert U_z \Vert_{\infty}+(1-z_{aj})
	   \Vert U_r \Vert_{\infty}(\frac{1}{4} \check \lambda \Vert U_f \Vert_{\infty}+\check \sigma_f), \\
	   \alpha_{\Delta_u} & = \frac{1}{4}(\check \lambda + \check \phi_r) \Vert W_z \Vert_{\infty}+(1-z_{aj})(\Vert W_r \Vert_{\infty}+\frac{1}{4} \check \lambda \Vert U_r \Vert_{\infty}\Vert W_f \Vert_{\infty}).
	\end{aligned}
	\end{equation}
Here $z_{aj}$ repectively $z_{bj}$ denote the latent state corresponding to $x_{aj}$ and $x_{bj}$.
It holds that 
\begin{equation}
\begin{aligned}
&\alpha_{\Delta_x}+\Vert U_o \Vert \alpha_{\Delta_u} \\
= & z_{aj}\left(1-\Vert U_r \Vert_{\infty}(\frac{1}{4}\check \lambda \Vert U_f \Vert_{\infty}+\check \sigma_f)-\Vert U_o \Vert_{\infty}(\Vert W_r \Vert_{\infty}+\frac{1}{4} \check \lambda \Vert U_r \Vert_{\infty} \Vert W_f \Vert_{\infty}) \right) \\
& +\frac{1}{4}(\check \lambda + \check \phi_r)\Vert U_z \Vert_{\infty}+\Vert U_r \Vert_{\infty}(\frac{1}{4} \check \lambda \Vert U_f \Vert_{\infty}+\check \sigma_f) \\
&+\Vert U_o \Vert_{\infty} \frac{1}{4}(\check \lambda + \check \phi_r) \Vert W_z \Vert_{\infty} + \Vert U_o \Vert_{\infty} (\Vert W_r \Vert_{\infty}+\frac{1}{4} \check \lambda \Vert U_r \Vert_{\infty}\Vert W_f \Vert_{\infty}).
\end{aligned}
\end{equation}
\end{proof}

Since $z_{aj} \in [1-\check \sigma_z, \check \sigma_z]$ this yields
\begin{equation}
\begin{aligned}
&\alpha_{\Delta_x}+\Vert U_o \Vert \alpha_{\Delta_u} \\
\leq &\check \sigma_z+\Vert U_r \Vert_{\infty}(\frac{1}{4} \check \lambda \Vert U_f \Vert_{\infty}+\check \sigma_f)(1-\check \sigma_z) +\Vert U_o \Vert_{\infty}(\Vert W_r \Vert_{\infty}+\frac{1}{4} \check \lambda \Vert U_r \Vert_{\infty} \Vert W_f \Vert_{\infty})(1-\check \sigma_z) \\
&+\Vert U_o \Vert_{\infty} \frac{1}{4}(\check \lambda + \check \phi_r) \Vert W_z \Vert_{\infty} \\
& \leq \check \sigma_z+(1-\sigma_z)\left(\Vert U_r \Vert_{\infty}(\frac{1}{4} \check \lambda \Vert U_f \Vert_{\infty}+\check \sigma_f)+\Vert U_o \Vert_{\infty}(\Vert W_r \Vert_{\infty}+\frac{1}{4} \check \lambda \Vert U_r \Vert_{\infty} \Vert W_f \Vert_{\infty}) \right) \\
&+\frac{1}{4}(\check \lambda + \check \phi_r)(\Vert U_z \Vert_{\infty}+\Vert W_z \Vert_{\infty} \Vert U_o \Vert_{\infty}).
\end{aligned}
\end{equation}

\subsection{GRU as an observer}
Lemma \ref{lem:GRUproperties} demonstrates that the GRU is a contraction under some conditions.
From there, it is easy to see that the GRU forgets its initial values. 
Note that we consider the rollouts after the warmup phase, where the GRU receives output feedback. 
This is done since we are interested in the long-term behavior of the system. 
During the warmup phase, the GRU receives the measumements as an input.
In this case, the setting reduces to the setting presented in \citet{BONASSI2021105049}.
This leads to weaker requirements for the system to be a contraction. 

\begin{theorem}[Observer GRU] \label{observer_GRU}
Consider a GRU \eqref{eq:GRUequation} and assume that the requirements in Lemma \ref{lem:GRUproperties} hold. 
Then, the GRU forgets is initial conditions.
Further, consider two rollouts $x$ and $\tilde x$ with initial values $x_a$ and $x_b$. 
Then it holds that 
\begin{equation}
\Vert \tilde x_n -x_n \Vert \rightarrow 0, n \rightarrow \infty. 
\end{equation}
\end{theorem}
	
\begin{proof}
Holds due to the contraction property.  
\end{proof}	

Theorem \ref{observer_GRU} shows that under certain conditions, the GRU dynamics forgets its initial conditions.
Intuitively, if the dynamics fit the data, it will behave as an observer via the simulator.
However, the GRU is not restricted to this properties.  
Further, there is no guarantee that the dynamics and observations can indeed be represented with GRU dynamics fulfilling the required properties.
In contrast, the existence is guaranteed for the KKL observer under mild assumptions. 
By design, the KKL also forgets its initial conditions. 
This can be seen by considering the transformation $T$ with Lipschitz constant $L$. It holds that 
\begin{equation}
\Vert T(Dz_n+B\hat s_n)-T(D\tilde z_n+B\hat s_n) \Vert \leq L \Vert D^n (z_0-\tilde z_0) \Vert \leq L |\lambda|^n\Vert z_0-\tilde z_0\Vert,
\end{equation}
where $\lambda$ denotes the maximum eigenvalue of $D$.

In the following, we will derive specific GRU architectures for special cases.
Theorem \ref{observer_GRU} provides some requirements, under which the GRU reconstructs the full latent state via the simulator. 
However, this is usually not the case that we consider in practice.
As derived in the main paper, we typically consider a partially OVS system with states $u$ that can be reconstructed via the simulator and states $v$ that can not. 
Therefore, we will demonstrate conditions under which a GRU can represent such a partially OVS system.
Intuitively, the GRU matrices are chosen, such that one part of the states is only influenced by the simulator and itself. 
After that we will derive the criteria, under which the GRU could act as an observer for the first part of the states.

\begin{lemma}[Partially OVS Representation] \label{lem:GRUSplit}
Consider the split of $x$ in $u$ and $v$ with $u \in \mathbb{R}^{d_u}, v \in \mathbb{R}^{d_x}$ and $d_x = d_u+d_v$. 
Further, consider GRU dynamics $f$ with matrices 
\begin{equation}
\begin{aligned}
W_r&=(0, W_r^v), \textrm{ with } W_r^v \in \mathbb{R}^{d_v \times d_y}, U_r = \begin{pmatrix}
U_r^u & 0 \\
U_r^{v,l} &  U_r^{v,r}
\end{pmatrix}, \\
W_z&=(0, W_z^v), \textrm{ with } W_z^v \in \mathbb{R}^{d_v \times d_y}, U_z= \begin{pmatrix}
U_z^u & 0 \\
U_z^{v,l} &  U_z^{v,r}
\end{pmatrix}, \\
W_f&=(0, W_f^v), \textrm{ with } W_f^v \in \mathbb{R}^{d_v \times d_y}, U_f= \begin{pmatrix}
U_f^u & 0 \\
U_f^{v,l} &  U_f^{v,r}.
\end{pmatrix}
\end{aligned}
\end{equation} 	
Then, the GRU dynamics $f$ with observation function $U_o$ can be split into $f_u$, $f_v$  
and observation functions $g$ and $r$, where 
\begin{equation}\label{eq:partially_obs}
\begin{aligned}
u_{n+1}&=f_u(u_n, i_n) \\
v_{n+1}&=f_v(u_n, v_n, \tilde y_n, i_n) \\ 
y_n & = g(u_n)+r(v_n). \\
\end{aligned}
\end{equation}
\end{lemma}
 \begin{proof}
 The matrices are constructed, such that $u$ is only influenced by $u$ and $i$ due to the choice of $W_r, W_z, W_u, U_r, U_z$ and $U_f$. The activation functions and Hadamard product are applied component-wise.  
 Note, that $f_u$ can be represented as GRU again and is thus a sub-GRU. 
 \end{proof}
\begin{lemma}[Partially OVS GRU]\label{lem:PartiallyOVSGRU}
	Consider the GRU \eqref{eq:GRUequation}, two rollouts $x$ and $\tilde x$ with initial values $x_a$ and $x_b$ and identical control inputs $i$. 
	If the partially OVS conditions from Lemma \ref{lem:GRUSplit} hold and further 
	\begin{equation}
	\Vert U_r^u \Vert_{\infty}\left(\frac{1}{4} \check \lambda \Vert U_f^u \Vert_{\infty} + \check \sigma_f\right)<1-\frac{1}{4} \frac{\check \lambda + \check \phi_r}{1-\check \sigma_z}
	\Vert U_z^u \Vert_{\infty}, \end{equation}
	with 
	\begin{equation}
	\begin{aligned}
	\check{\sigma}_z &=\sigma(\Vert \tilde W_z^u \ \check \lambda U_z^u \ b_z^u \Vert_{\infty}) \\
	\check{\sigma}_f & = \sigma(\Vert \tilde W_f^u \ \check \lambda U_f^u \ b_f^u \Vert_{\infty}) \\
	\check{\phi}_r &= \phi(\Vert \tilde W_r^u \ \check \lambda U_r^u \ b_r^u) \Vert_{\infty}),
	\end{aligned}
	\end{equation}
	where the superscript $u$ denotes the submatrices that affect $u$ 
	then it holds that 
	\begin{equation}
\Vert u_n-\tilde u_n \Vert \rightarrow 0
	\end{equation}	
\end{lemma} 
\begin{proof}
Since the matrices are constructed such that $u$ forms an independent system that is not influenced by $y$ and the other part of the latent states $v$, the proof of Theorem 2 in \citet{BONASSI2021105049} can be applied. In particular consider $\Delta u^+ = f_u(u_a,i_0)-f_u(u_b,i_0)$ and $\Delta u = u_a-u_b$. 
Then it holds that
\begin{equation}
\Vert \Delta u^+ \Vert_{\infty} \leq (1-\delta)\Vert \Delta u \Vert_{\infty},
\end{equation}
with $\delta \in (0,1)$. 
Thus, the system is a contraction and it holds that 
\begin{equation}
\Vert u_n-\tilde u_n \Vert \rightarrow 0.
\end{equation}
\end{proof}

The results show that the GRU architectur can represent a split in $u$ and $v$, where the transitions of $u$ form a sub-GRU.
Thus, under some requirements, the sub-GRU is a contraction for $u$ and thus, $u$ forgets its initial conditions. 
This means that if we can reproduce the original OVS system with this architecture, then the sub-GRU is able to learn a system, where $u$ can be reconstructed via the simulator. 
We expect that it in this cases it can also balance the small model mismatch, continuously correcting the latent states $u$ similar to the KKL-RNN.
However, as before, there is no guarantee that a required system exists for all data.
Further, the required properties are not enforced in the GRU architecture by design. 
Also, the desired split is not enforced in cotrast to our KKL-RNN architecture. 
We will show later that this could lead to a model that ignores the simulator. 
For completeness, we will first show how to obtain the fully OVS case. 
 
\paragraph{Fully OVS case: }
The fully OVS case is obtained by ignoring the data input, thus $W_r=0$, $W_z=0$, and $W_f=0$.
In this case, the system reduced to a system with control input $i$ and without output feedback. 
Thus, if the conditions from \citet{BONASSI2021105049} hold, the GRU is a contraction and can behave as an observer. 
Again, this requires that the original system can be represented with such a GRU architecture.
This includes that the original system is OVS. 

\paragraph{Ignoring the simulator: }
In any case, it is also possible to construct a dynamical system that is able to predict the data $y$ without considering the simulator inputs $\hat s$. 
Intuitively, this holds since the data are observable via the data itself.
Thus, latent dynamics can be constructed explaining the data solely via the data.
This can be easily seen by considering $\tilde W_r=0$, $\tilde W_z=0$, and $\tilde W_f=0$, thus ignoring the simulator input.
By ignoring the simulator it is visible easily that the acthitecture can not inform the latent states via the simulator.
However, this is also possible in other scenarios if the system violates the observer properties and allows errors to accumulate. 

\subsection{Summary and interpreting the results}
In the experiments, we have seen that the hybrid GRU acts similarly to our KKL-RNN in some cases.
Indeed, under certain conditions the GRU dynamics are a contraction and thus, the GRU forgets its initial condition.
The GRU architecture further allows to split the internal latent states into an OVS and non-OVS part, where the OVS part is solely addressed by the simulator and not by output feedback. 
Thus, if the dynamics additionally match the data, the GRU could act similar as an observer. 
In this cases, the setting allows to inform the OVS latent states via the simulator as before and we expect it to balance small modeling mismatchtes, thus correcting and stabilizing the predictions similar to our KKL-RNN. 
However, there is no formal proof that such a GRU exists for all system. 
Further, the necessary properties for a contraction are not fulfilled by design. 
We further showed that it is also possible to ignore the simulator inputs. 
In the experiments we observed, that the KKL-RNN was especially beneficial in case the simulator is only partially informative, e.g. in the partially OVS case. 
It can be interpreted that it is easier for the GRU to act as an observer if the simulator data are easy to process and no split in OVS and non-OVS has to be learned.
To interpret the results further, consider a fully OVS system. 
The hybrid GRU constantly receives the simulator input.
In contrast, the data are only provided during a warmup phase and are further noisy.
Thus, it might be easier to inform the latent states via the simulator than via the data.
The partially OVS system on the other hand requires learning the correct split, making the learning-task harder again.  
Additionally, if the system can not easily detect the states that are OVS via the simulator, this could also cause problems.
Later we will demonstrate experimentally that also GRUs specifically trained on the partially OVS systems are not able to learn the correct split.

\newpage
\newpage
\section{Additional results and experiments} \label{section:add}
In this section, we will provide additional results.
We will first complete the results from the experiments section by adding plots, RMSE over time and runtimes. 
To further analyze the method, we will add an ablation study based on the partially OVS system presented in the problem formulation. 
Further, we will demonstrate how our method can be extended to invertible neural networks, allowing to directly address the dynamics $f_u$. 

\subsection{System v): Pure learning-based scneario}
We present an additional example for the pure learning-based scenario and refer to it as system (v).
We extend the experiment in \citep{ensinger2023combining} Exp. ii) and use identical data from the double-torsion pendulum \citep{lisowski} with a varied control input.
As in their setting, we learn a simulator substitute by training a GRU on the low-pass filtered and downsampled signal. 
We refer to their method as Filter. 
However in contrast to their setting, we inform the high-pass via the low-pass components. 
This is done by feeding them as an input to our KKL-RNN and the hybrid GRU.
The results demonstrate that we can further improve the concept in \citet{ensinger2023combining}.
While the GRU shows accumulating errors, Residual model and Filter both benefit from the stable long-term behavior of the low-pass component.
However, since low-pass and high-pass component are not linked for those models, the high-pass component still suffers from accumulating errors and deteriorated behavior, which is prevented by our KKL-RNN and the hybrid GRU.
This finding is more clearly visible in the rollout plots \ref{subfig:4_baseline} than in the RMSE \ref{t:lb} since it mainly refers to the high-frequency components. 
The signal however is mainly dominated by the low frequency behavior that Residual Model, Filter, hybrid GRU and hybrid KKL-RNN share.

\begin{table*}[h!]
	\centering 	
	\caption{Total RMSEs for systems iv)-v) (mean (std)) over 5 independent runs.}
	\label{t:lb}
	\begin{tabular}{rccccc}
		\noalign{\smallskip} \hline    \noalign{\smallskip}
		task & GRU & Residual Model & Filter &Hybrid GRU & Hybrid KKL-RNN (ours)  \\
		\hline
		v) & 1.18 (0.29) & 0.64 (0.07) & 0.33 (0.06) & \textbf{0.29} (0.07) & \textbf{0.29} (0.09)  \\
		\noalign{\smallskip} \hline \noalign{\smallskip}
	\end{tabular}\quad
\end{table*} 

\subsection{GRU ablation study}
As a recap, our proposed KKL-RNN architecture consists of
\begin{itemize}
	\item Partially observable system containing OVS and non-OVS latent states; 
	\item Specific loss function penalizing the non-OVS components;
	\item KKL-observer. 
\end{itemize}
The goal of this study is to analyze how different parts of the architecture affect the results.
In particular, we try to find out how the split in OVS and non-OVS affects the results and how the KKL observer affects the results in contrast to an architecture built from different GRUs.  
To this end, we consider three GRU architectures to further analyze the system based on the partially OVS system presented in Section 2
with $f_u:\mathbb{R}^{d_u} \times \mathbb{R}^{d_s} \rightarrow \mathbb{R}^{d_u}$ and $f_v:\mathbb{R}^{d_u} \times \mathbb{R}^{d_v} \rightarrow \mathbb{R}^{d_v}$.
Next, we consider single components of the partially OVS system or the whole system.
We model all components with GRUs. 
In the first scenario, we extend the hybrid GRU with an observation model $h$ that aims to reproduce the simulator. 
The transitions remain the same as in the hybrid GRU setting. 
With this experiment we analyze if it is sufficient to simultanously model the simulator. 
This forces the model to learn latent states that are able to reproduce the simulator. 
For the remaining architectures, we aim to reproduce the dynamics in the partially OVS systems.
To this end, we will train separate GRUs on $f_u$ and $f_v$. 
As before, $g$, $r$ and $h$ will be modeled with linear layers. 
We vary the input the GRUs receive. 
To make the architecture comparable to the hybrid KKL-RNN, we will apply the same losses to Scenario 2 and Scenario 3.
Providing the observations as an input to both GRUs in Scenario 2), we analyze whether it is sufficient to consider the proposed split and the corresponding loss function.
In the third scenario, we aim to reproduce our KKL-RNN architecture as closely as possible.
This is done by providing only the simulator as an input to the first model and only the data as an input to the second model. 
With this experiment, we study, whether the KKL observer could also be replaced with an additional GRU receiving identical input data. 
In summary, we consider the following architectures. 
\begin{itemize}
	\item Scenario 1: We extend the hybrid GRU with an additional observation function $h$ modeling the simulator. 
	\item Scenario 2: $f_u$ and $f_v$ are modeled with standard GRUs receiving output feedback. I.e., both receive the data and later output feedback as a control input. 
	\item Scenario 3: $f_u$ receives only the simulator as control input, $f_v$ receives the simulator and output feedback as control input. This mimics the setting of our KKL-RNN.  
\end{itemize}

\subsubsection{Results} 
\begin{figure*}[ht!]
	\begin{subfigure}{0.49\textwidth}
		\centering
		\includegraphics[width= \textwidth]{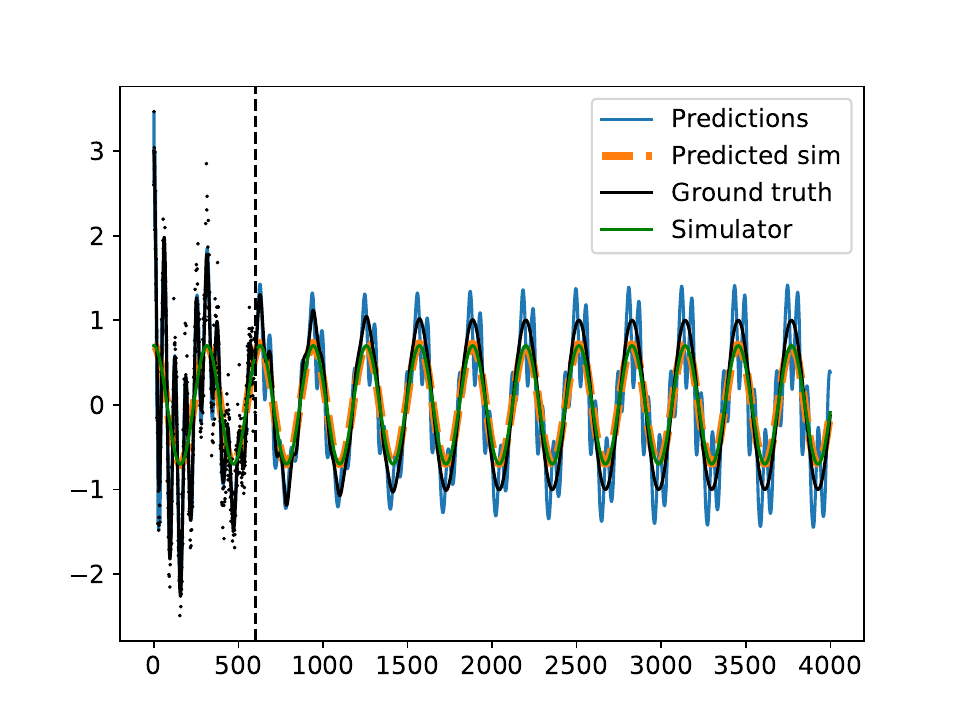}
		\caption{Scenario 1 on system i)} \label{subfig:S1mass}
	\end{subfigure}	
	\begin{subfigure}{0.49\textwidth}
		\centering
		\includegraphics[width= \textwidth]{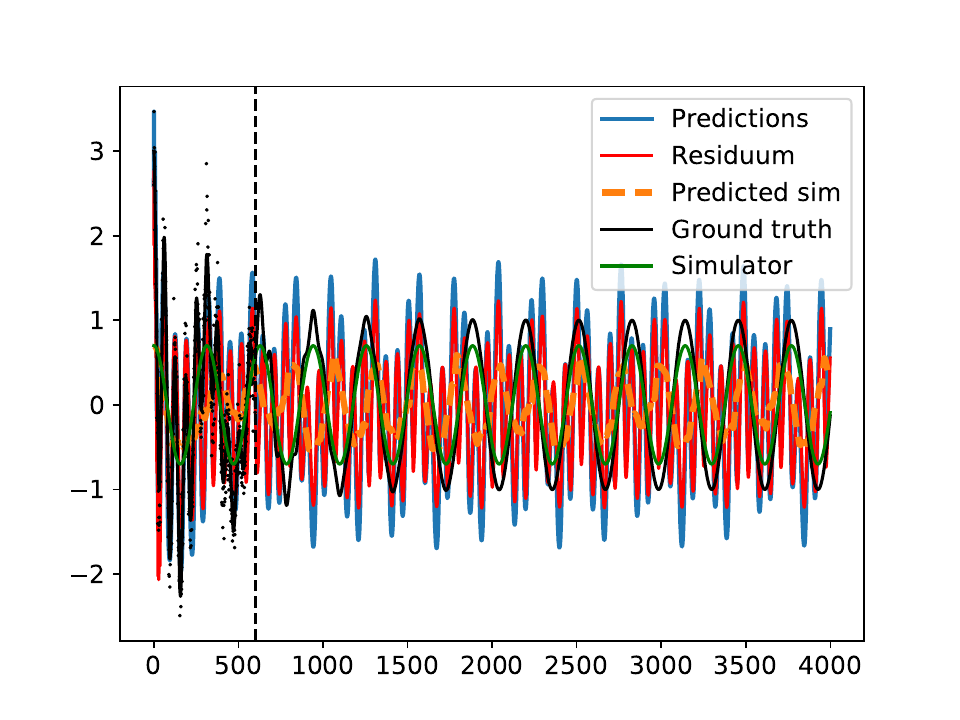}
		\caption{Scenario 2 on system i)} \label{subfig:S2mass}
	\end{subfigure}	
	\begin{subfigure}{0.49\textwidth}
		\centering
		\includegraphics[width= \textwidth]{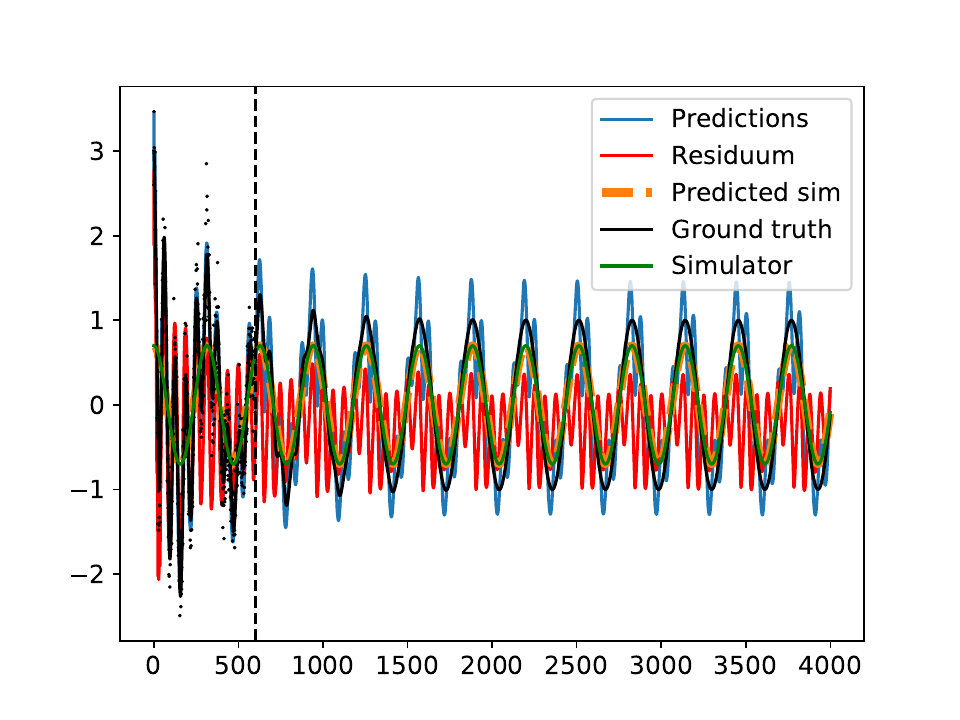}
		\caption{Scenario 3 on system iii)} \label{subfig:S3mass}
	\end{subfigure}	
	\caption{Plots for system i). Shown are the results for Scenario 1 in Fig. \ref{subfig:S1mass}, Scenario 2 in Fig. \ref{subfig:S2mass} and Scenario 3 in Fig. \ref{subfig:S3mass}. All models have problems to reproduce the oscillations and Scenario 1 even shows upswinging oscillations. Further, the models are not able to learn the correct split in OVS and non-OVS. For Scenario 3 it is clearly visible that the residuum takes over most of the prediction task.}
	\label{fig:mass}
\end{figure*}

\begin{figure*}[ht!]
	\begin{subfigure}{0.49\textwidth}
		\centering
		\includegraphics[width= \textwidth]{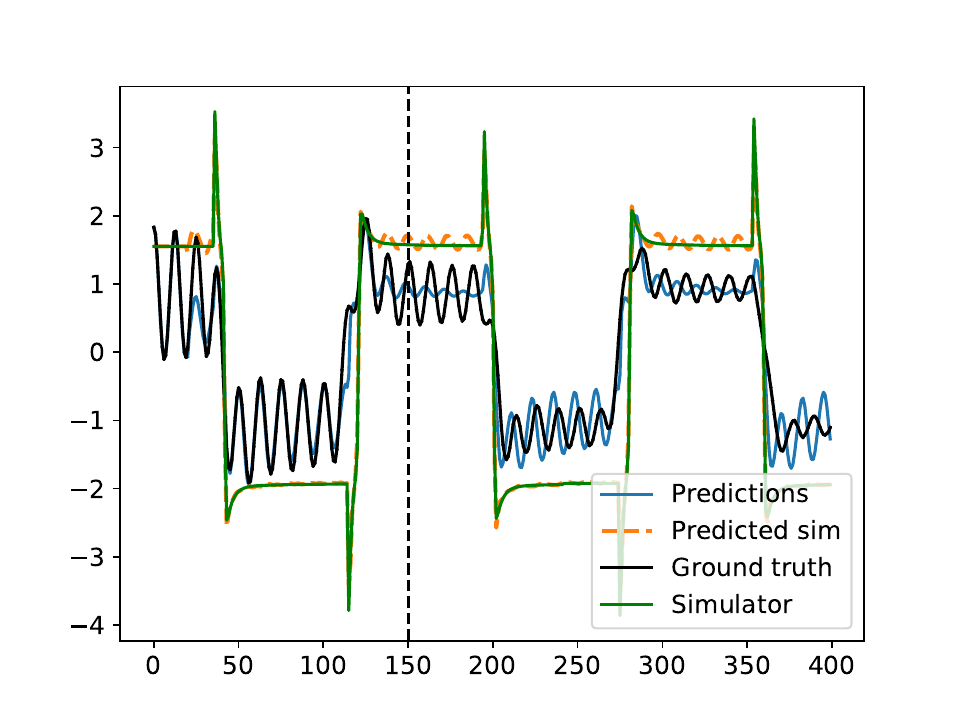}
		\caption{Scenario 1 on system ii)} \label{subfig:S1DT}
	\end{subfigure}	
	\begin{subfigure}{0.49\textwidth}
		\centering
		\includegraphics[width= \textwidth]{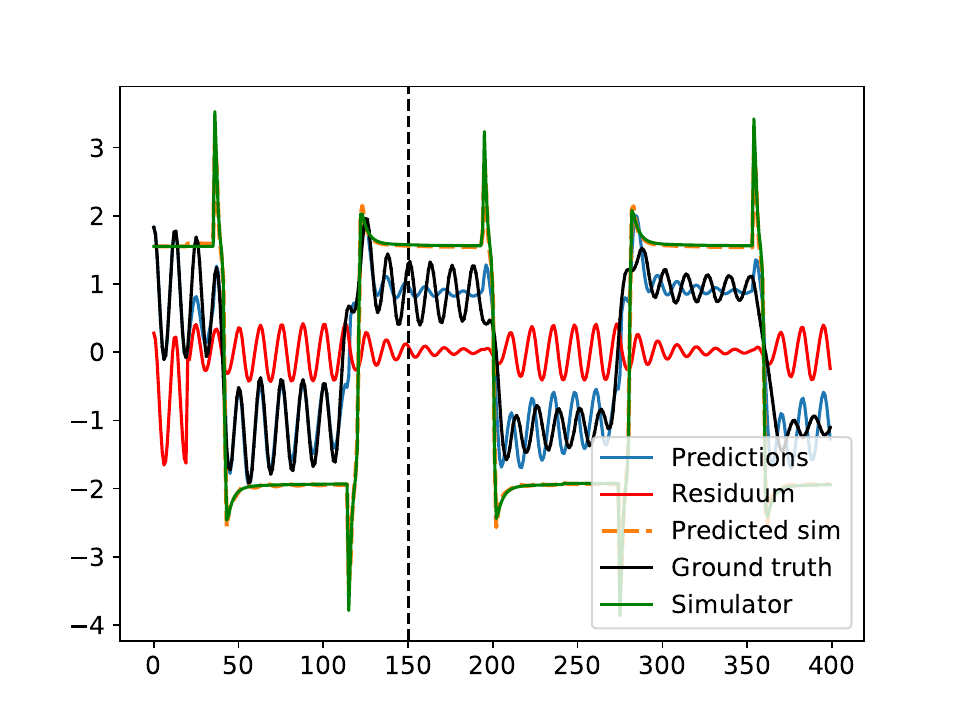}
		\caption{Scenario 3 on system ii)} \label{subfig:S2DT}
	\end{subfigure}	
	\caption{Plots for system ii). Shown are the results for Scenario 1 in Fig. \ref{subfig:S1DT} and Scenario 3 in Fig. \ref{subfig:S2DT}.
		Both methods are not able to reproduce the oscillations correctly and Scenario 3 learns a wrong split in OVS and non-OVS.}
	\label{fig:mass}
\end{figure*}

\begin{figure*}[ht!]
	\begin{subfigure}{0.49\textwidth}
		\centering
		\includegraphics[width= \textwidth]{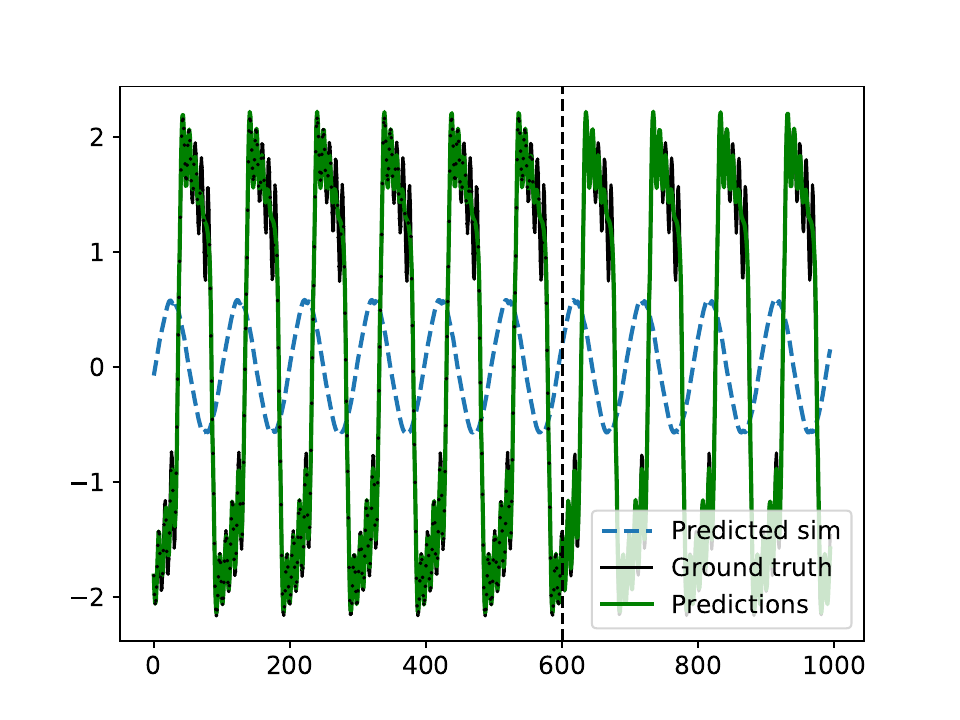}
		\caption{Scenario 1 on system iv)} \label{subfig:S1VDP}
	\end{subfigure}	
	\begin{subfigure}{0.49\textwidth}
		\centering
		\includegraphics[width= \textwidth]{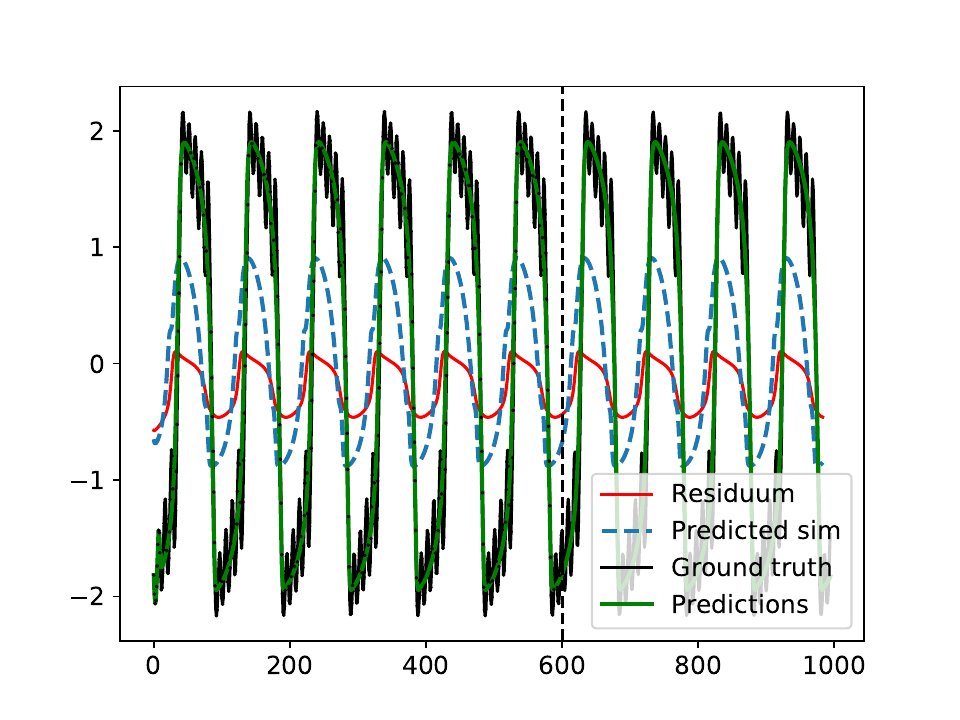}
		\caption{Scenario 2 on system iv)} \label{subfig:S2VDP}
	\end{subfigure}	
	\begin{subfigure}{0.49\textwidth}
		\centering
		\includegraphics[width= \textwidth]{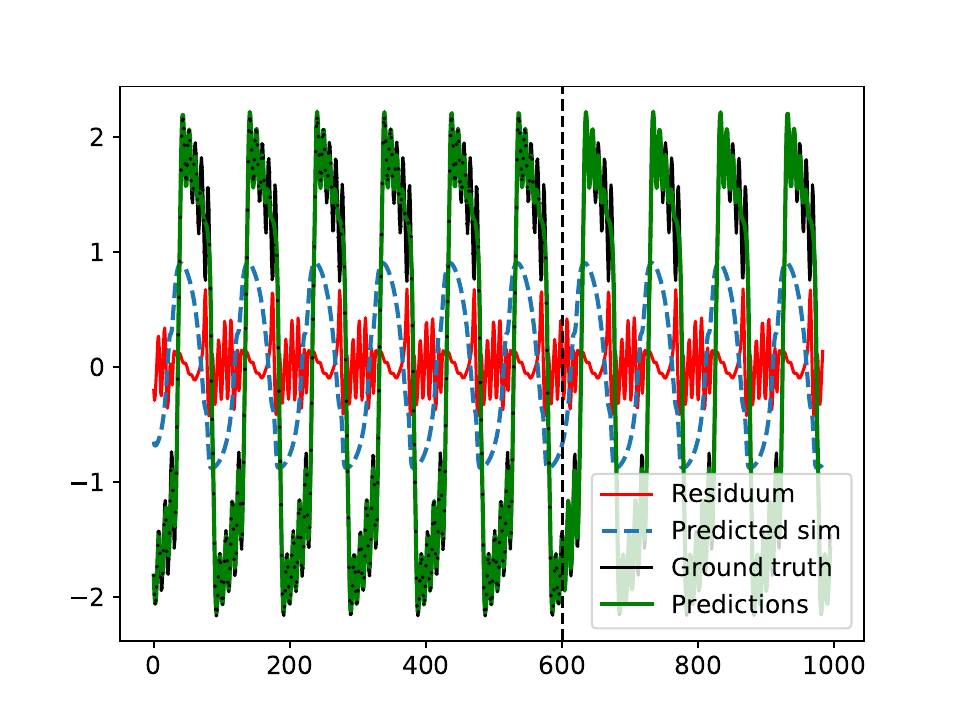}
		\caption{Scenario 3 on system iv)} \label{subfig:VDP}
	\end{subfigure}	
	\caption{Plots for system i). Shown are the results for Scenario 1) in Fig. \ref{subfig:S1VDP}, Scenario 2) in Fig. \ref{subfig:S2VDP} and Scenario 3) in Fig. \ref{subfig:VDP}. Again, the models learn a wrong split and are not able to reproduce the oscillations correctly.}
	\label{fig:mass}
\end{figure*}

The results are presented in Table \ref{t:ablation}. 
They indicate that Scenario 2 is not enough to model the whole system, probably because the data do not contain enough information to model also the simulator for all systems. 
This is for example demonstrated in Fig. \ref{subfig:S2mass}.
In contrast, this information is provided in Scenario 1 and 3 by providing the simulator as control input to both models. 
It can be seen that Scenario 1 has a similar accuracy as the hybrid GRU and suffers from similar problems. 
Especially, it also shows the deteriorated behavior in system i) (cf. Fig. \ref{subfig:S1mass}) and can not reproduce the oscillations for system ii) 
(cf. Fig \ref{subfig:S1DT}) and iv) (cf. Fig. \ref{subfig:S1VDP}). 
In contrast to Scenario 1, Scenario 3 is informed about the split in OVS and non-OVS part in its architecture similar to the KKL-RNN.
However, still Scenario 3 has similar problems with learning the correct oscillations. 
Further, it is also visible that it does not learn the correct split in OVS and non-OVS (cf. Fig. \ref{subfig:S3mass}, Fig. \ref{subfig:S2DT} and Fig. \ref{subfig:VDP}). 
The results indicate that the KKL-observer is a central component of our method and the split into the partially OVS system not enough.  
At the same time, the results show that the highest accuracy is in general achieved with Scenario 3, indicating that the proposed architecture is already useful in itself. 
This includes the split in OVS and non-OVS, the loss and an observer-inspired setup.  
The proposed architecture could thus also be applied in combination with standard recurrent networks. 

\begin{table*}[h!]
	\centering 
	\caption{Results for Scenario 1)-3) on systems i)-v) (mean (std)) over 5 indep. runs.}
	\label{t:ablation}
	\begin{tabular}{rccc}
		\noalign{\smallskip} \hline \noalign{\smallskip}
		System & Scenario 1 & Scenario 2  & Scenario 3   \\
		\hline
		i) &  \textbf{0.25} (0.12) & 0.97 (0.14) & \textbf{0.25} (0.07) \\
		ii)& 0.54 (0.29)  & 1.28 (0.03) & \textbf{0.41} (0.02) \\
		iii) & \textbf{0.24} (0.02) & 0.47 (0.12) & 0.27 (0.01) \\
		iv) & 0.1 (0.08) & 0.63 (0.43) & \textbf{0.06} (0.04)  \\
		v) & \textbf{0.28} (0.09) & 1.35 (0.24) & \textbf{0.28} (0.08) \\
		\noalign{\smallskip} \hline \noalign{\smallskip}
	\end{tabular}\quad
\end{table*}

\subsection{Invertible neural network} 
As described in Section 4, we provide the option to train with invertible neural networks in case direct access to the dynamics $f_u$ is required, e.g. to analyze for fixed points, include symmetries etc.
In detail, we learn $T_{\theta}^{\star}$ as invertible network allowing to access $T_{\theta}$ by inversion. 
An invertible network is obtained by stacking coupling layers \citep{DBLP:conf/iclr/DinhSB17}. 
For $f_u$ it holds that 
\begin{equation}
f_u=T^{\star}_{\theta}(D_{\theta}T_{\theta}u_n+F\hat{s}_n). 
\end{equation}
This allows to obtain $u_{n+1}$ directly via $f_u(u_n)$. 
Later, in Sec. \ref{sec:KKL}, we provide details on the architecture. 
We train the architecture on system i) for 5 independent random seeds.
This provides similar results as the default scenario containing a standard MLP. 
Fig. \ref{subfig:mass} shows a plot of the rollouts demonstrating that also the qualitative results are similar to the default setting.
In particular, the split in OVS and non-OVS is accurate and the simulator can be reconstructed.

\subsection{Additional plots} 
\begin{figure*}[h!]
	\begin{subfigure}{0.49\textwidth}
		\centering
		\includegraphics[width= \textwidth]{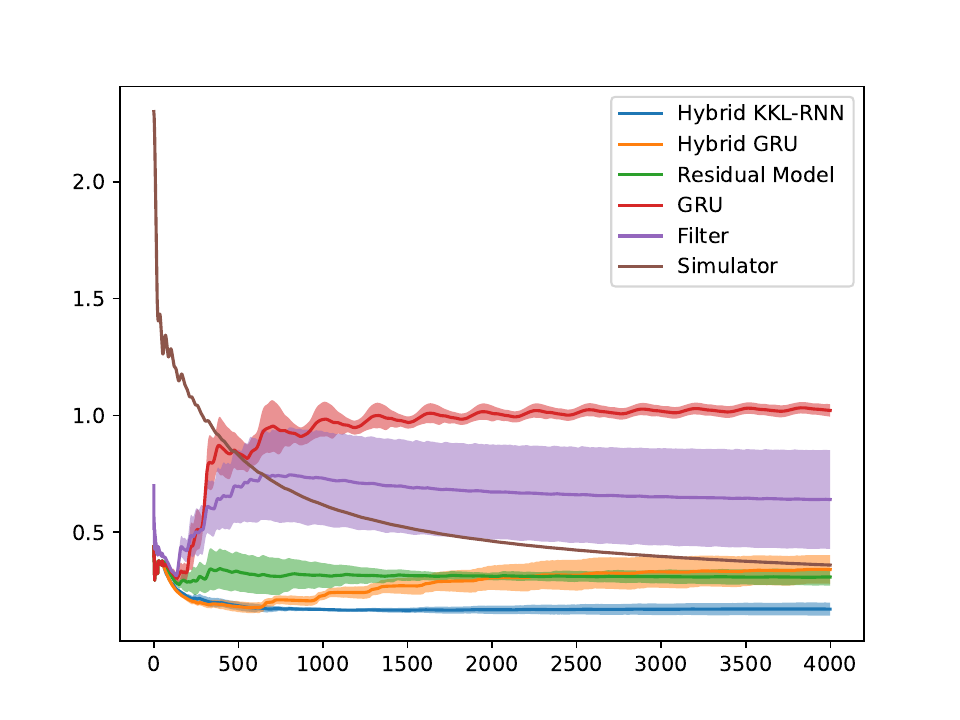}
		\caption{System i), RMSE over time} \label{subfig:RMSE_mass}
	\end{subfigure}	
	\begin{subfigure}{0.49\textwidth}
		\centering
		\includegraphics[width= \textwidth]{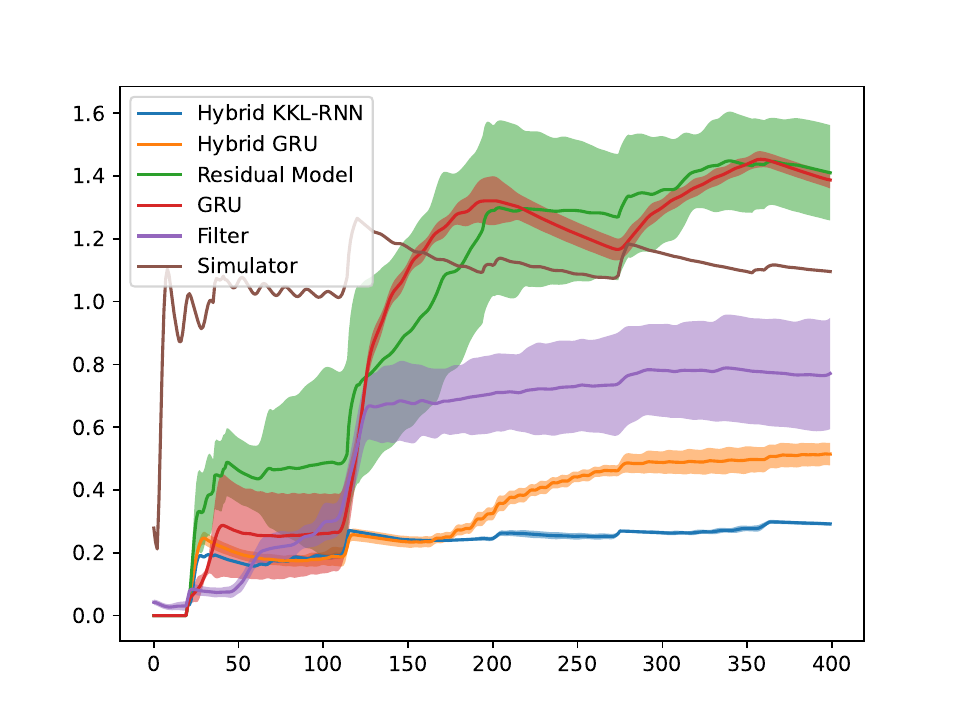}
		\caption{System ii), RMSE over time} \label{subfig:RMSE_double_torsion}
	\end{subfigure}	
	\begin{subfigure}{0.49\textwidth}
	\centering
	\includegraphics[width= \textwidth]{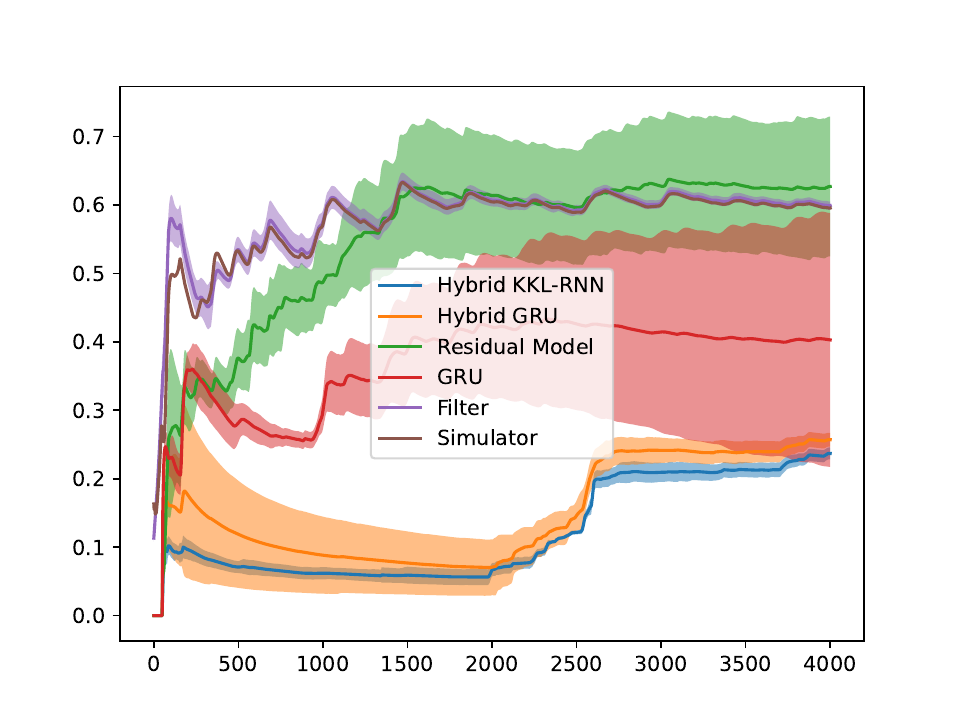}
	\caption{System iii), RMSE over time} \label{subfig:RMSE_friction}
\end{subfigure}	
	\caption{Accumulated RMSEs over time for systems i)-iii). Fig. \ref{subfig:RMSE_mass} shows the results for system i), 
		Fig. \ref{subfig:RMSE_double_torsion} shows the results for system ii) and Fig. \ref{subfig:RMSE_friction} shows the results for system iii).}
	\label{fig:hybrid}
\end{figure*}

\begin{figure*}[h!]
	\begin{subfigure}{0.49\textwidth}
		\centering
		\includegraphics[width= \textwidth]{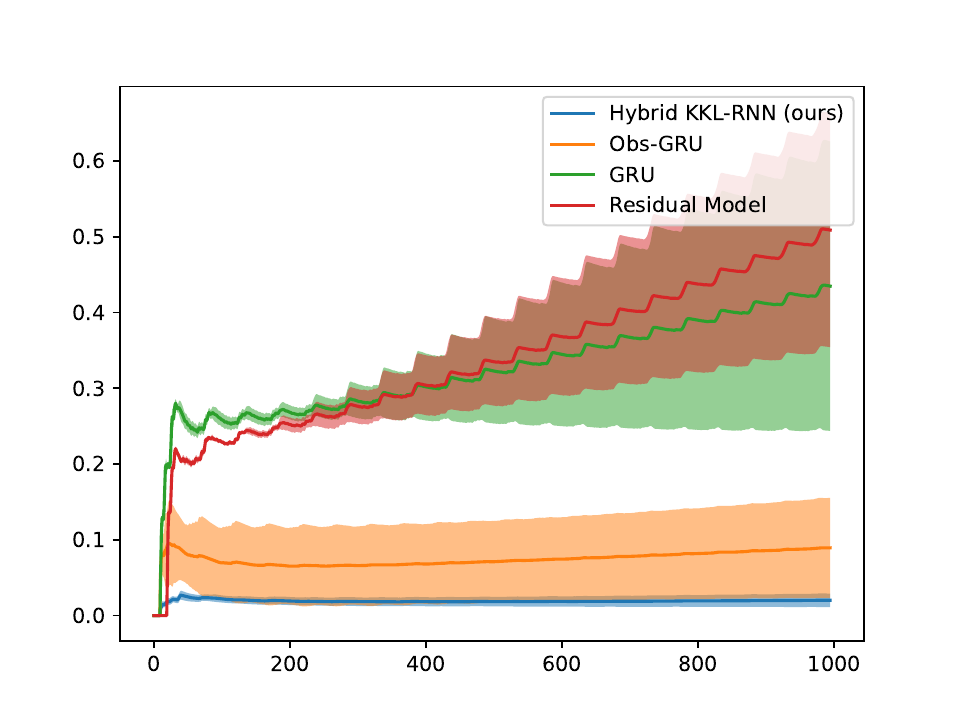}
		\caption{System iv), RMSE over time} \label{subfig:RMSE_VDP}
	\end{subfigure}	
	\begin{subfigure}{0.49\textwidth}
		\centering
		\includegraphics[width= \textwidth]{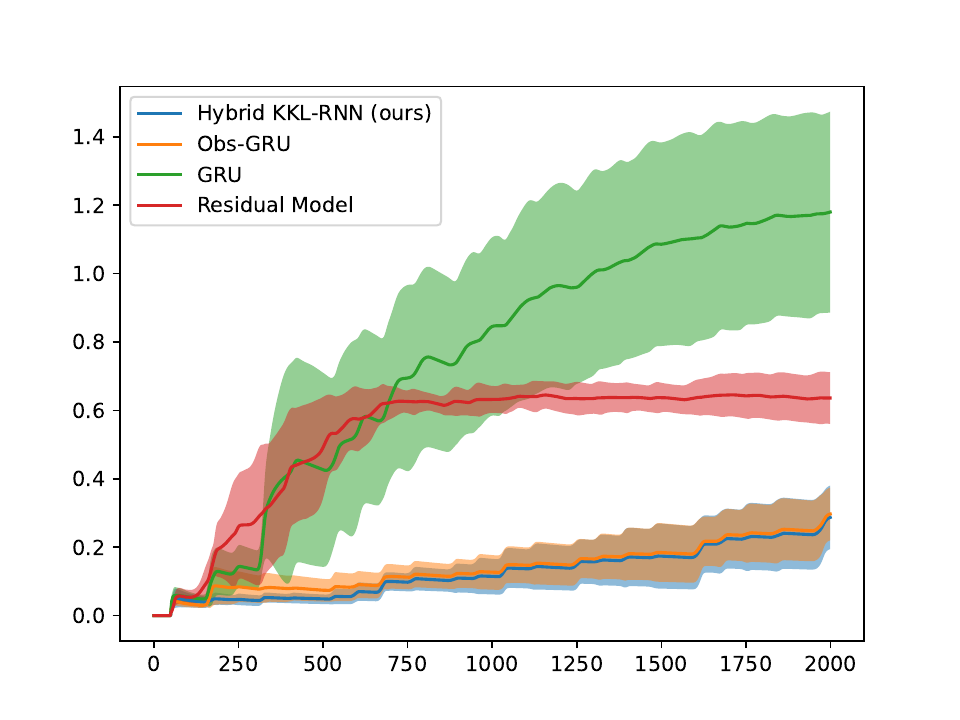}
		\caption{System v), RMSE over time} \label{subfig:RMSE_sim}
	\end{subfigure}	
	\caption{Accumulated RMSEs over time for systems iv) and v). Fig. \ref{subfig:RMSE_VDP} shows the results for system iv) 
  and Fig. \ref{subfig:RMSE_double_torsion} shows the results for system ii), \ref{subfig:RMSE_sim} shows the results for system v).}
	\label{fig:baselines}
\end{figure*}

\begin{figure*}[h!]
	\begin{subfigure}{0.49\textwidth}
		\centering
		\includegraphics[width= \textwidth]{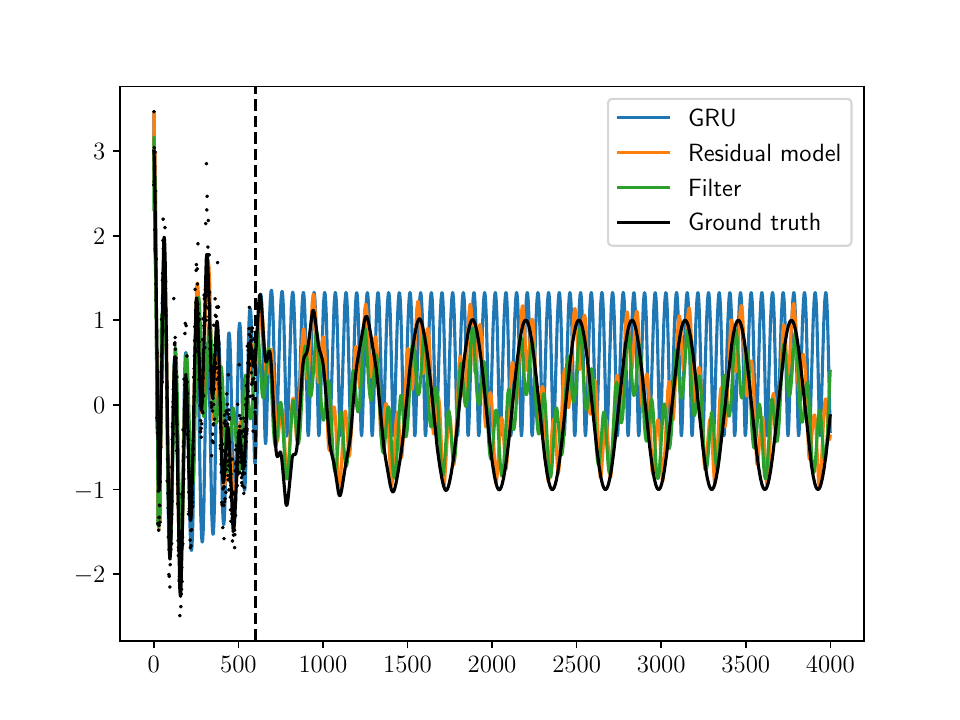}
		\caption{System i), baselines} \label{subfig:1_baselines}
	\end{subfigure}	
	\begin{subfigure}{0.49\textwidth}
		\centering
		\includegraphics[width= \textwidth]{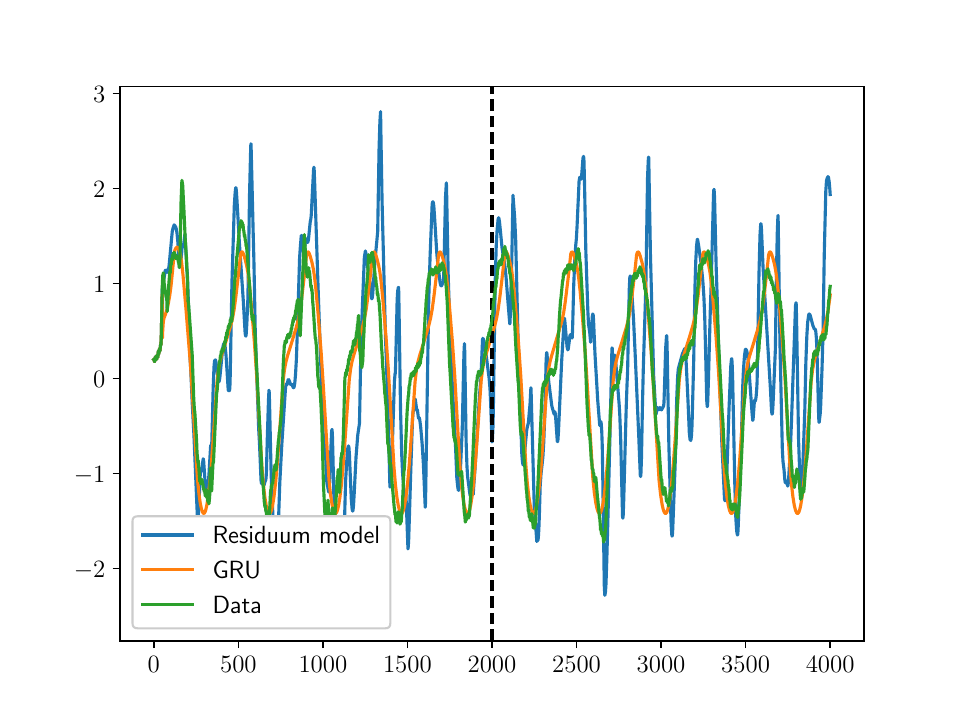}
		\caption{System iv), baselines} \label{subfig:4_baselines}
	\end{subfigure}	
	\caption{Baselines for system i) in Fig. \ref{subfig:1_baselines} and system iii) in Fig. \ref{subfig:4_baselines}. The baselines do not provide a good representation of the rollouts in both cases. Especially, they show deteriorated long-term behavior.}
	\label{fig:hybrid_KKL}
\end{figure*}

\begin{figure*}
	\begin{subfigure}{0.49\textwidth}
	\centering
	\includegraphics[width= \textwidth]{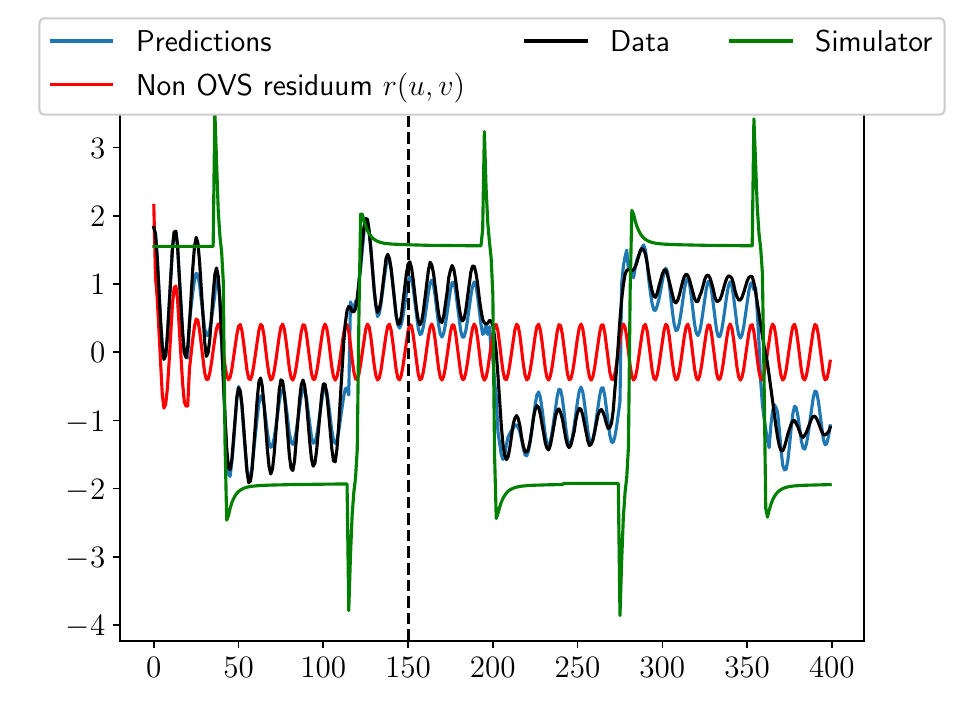}
	\caption{System i), baselines} \label{subfig:3_KKLRNN}
\end{subfigure}	
\begin{subfigure}{0.49\textwidth}
	\centering
	\includegraphics[width= \textwidth]{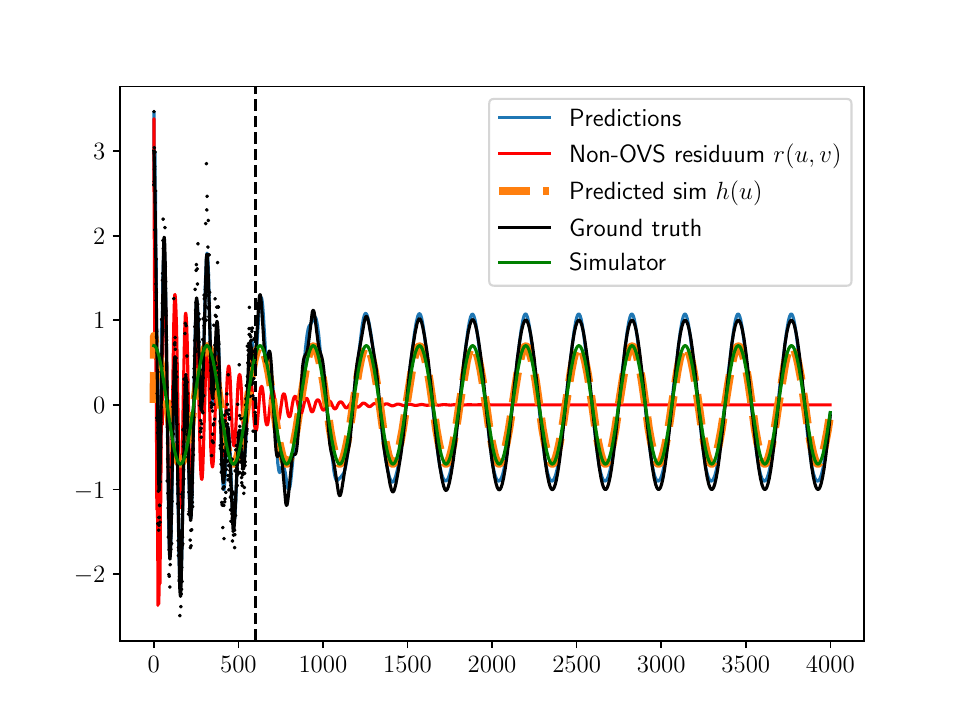}
	\caption{System i) with invertible neural network.} \label{subfig:mass}
\end{subfigure}
	\caption{Hybrid KKL-RNN for system 3) in Fig. \ref{subfig:3_KKLRNN}. The model learns the correct split in OVS and non-OVS. The oscillations are not perfectly reproduced but still a good representation. Fig. \ref{subfig:4_baselines} depicts system i) with the hybrid KKL-RNN and invertible transformation.}
\label{fig:hybrid_KKL_new}
\end{figure*}

\begin{figure*}[h!]
		\begin{subfigure}{0.49\textwidth}
			\centering
			\includegraphics[width= \textwidth]{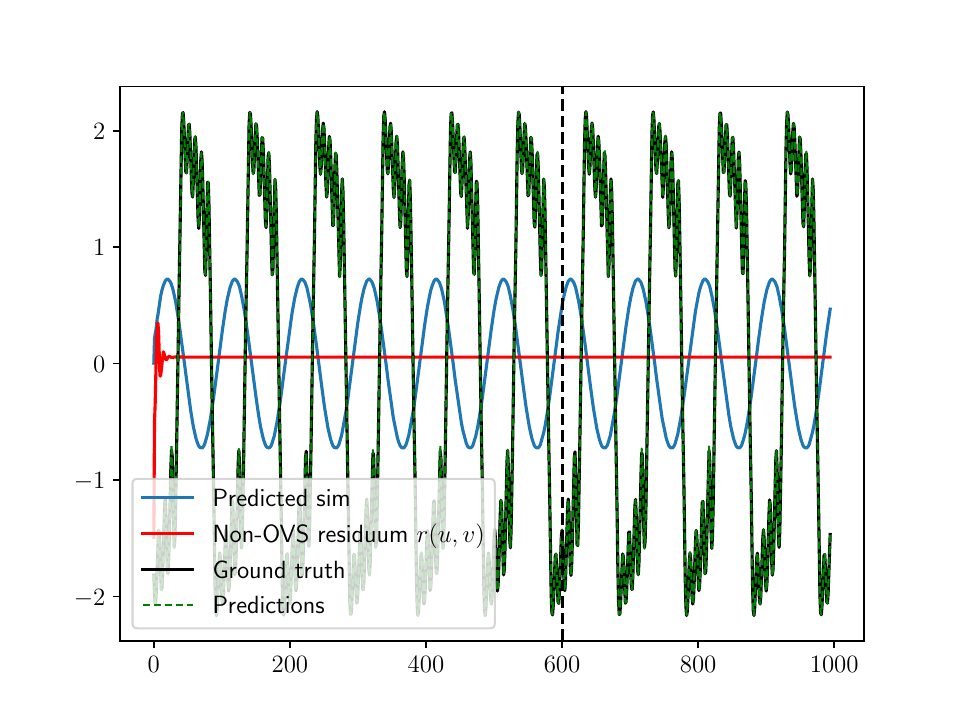}
			\caption{System iv), KKL-RNN} \label{subfig:6_baselines}
		\end{subfigure}	
		\begin{subfigure}{0.49\textwidth}
		\centering
		\includegraphics[width= \textwidth]{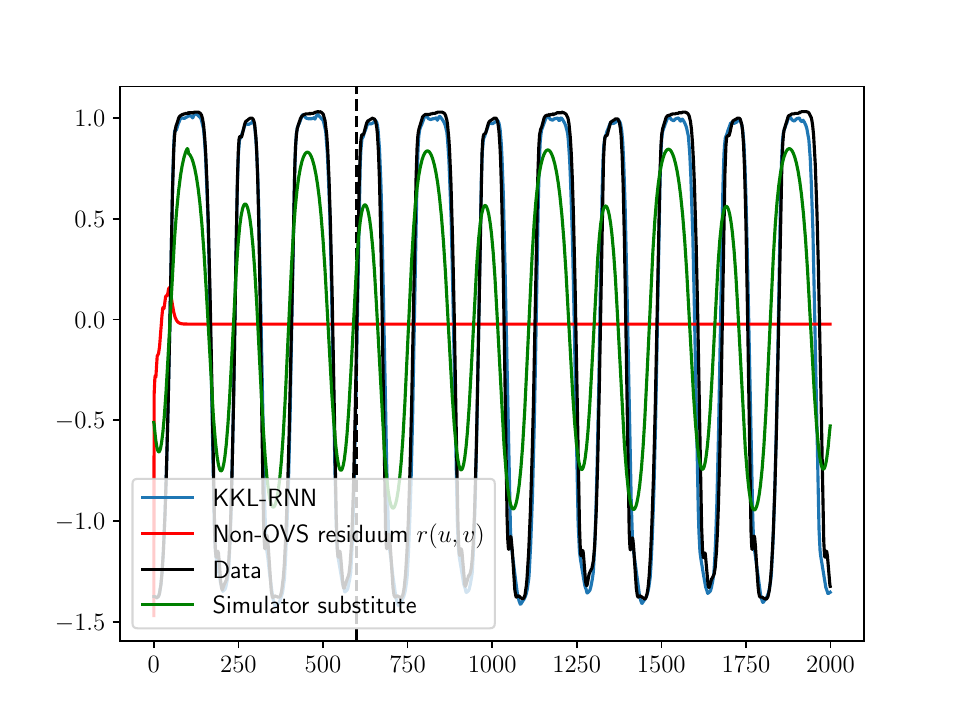}
		\caption{System v), KKL-RNN} \label{subfig:5_KKLRMM}
	\end{subfigure}	
		\caption{KKL-RNN for system iv) in Fig. \ref{subfig:6_baselines} and system v) in Fig. \ref{subfig:5_KKLRMM}.
		In both cases, the KKL-RNN learns the correct split in OVS and non-OVS and reproduces the predictions accurately.}
		\label{fig:lb}	
\end{figure*}

\begin{figure*}[h!]
	\begin{subfigure}{0.49\textwidth}
		\centering
		\includegraphics[width= \textwidth]{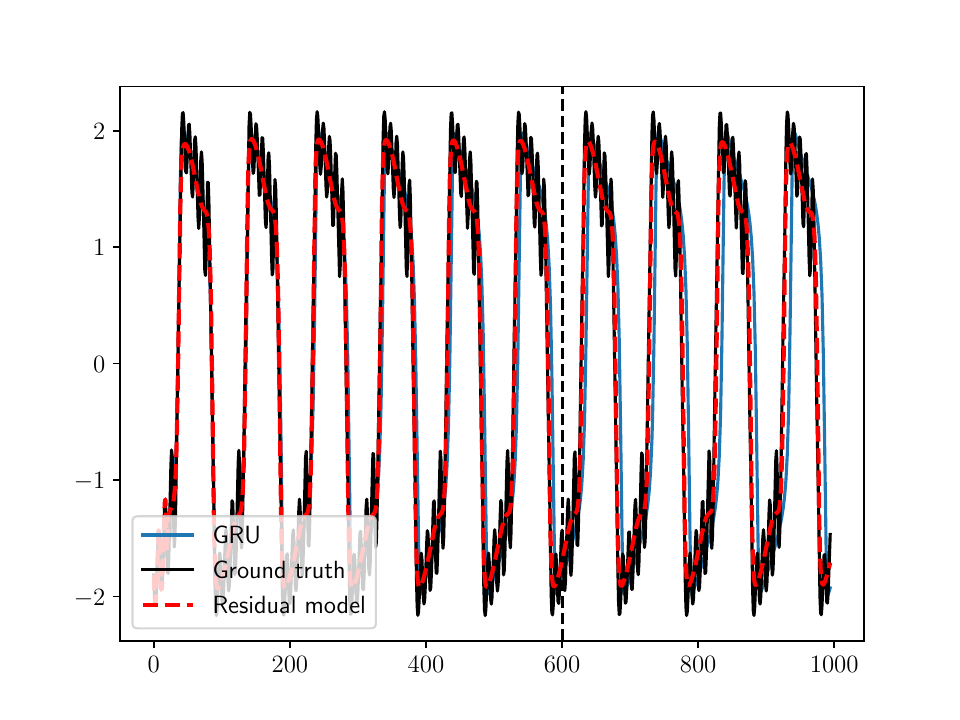}
		\caption{System iv), baselines} \label{subfig:6_baseline}
	\end{subfigure}	
	\begin{subfigure}{0.49\textwidth}
		\centering
		\includegraphics[width= \textwidth]{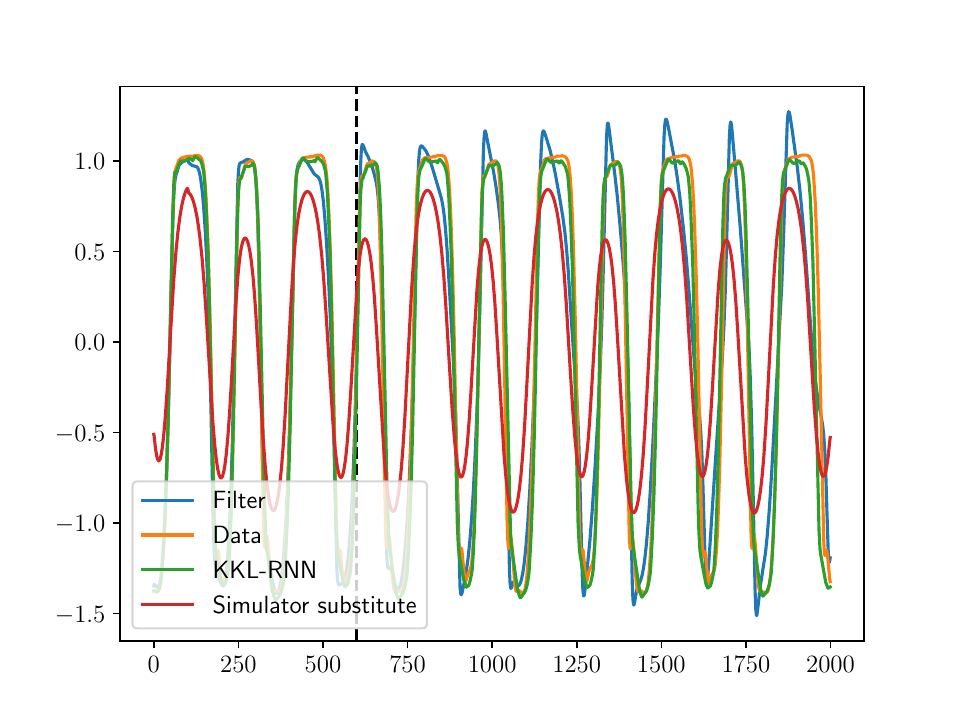}
		\caption{System v), baselines} \label{subfig:4_baseline}
	\end{subfigure}	
	\caption{Baselines for system iv) in Fig. \ref{subfig:6_baseline} and system v) in Fig. \ref{subfig:4_baseline}.
	The GRU suffers from the typical drift in system iv), while the residual model doesn't learn the oscillations correctly.
    For system v), the baselines show deteriorated behavior in the high-pass components.}
	\label{fig:lb_baselines}
\end{figure*}

We add missing plots for all experiments in the experimental section. 
In Fig. \ref{fig:hybrid}, we report the accumulated RMSEs over time for systems i-iii).
In Fig. \ref{fig:baselines}, we report the accumulated RMSEs over time for systems iv) and v). 
We provide plots for the baselines on systems i) and iii) in Fig. \ref{fig:hybrid_KKL}, indicating that they are not able to reproduce the dynamics well.
Fig. \ref{subfig:3_KKLRNN} shows the results of our hybrid KKL-RNN for system ii). 
It indicates that it reproduces the dynamics well even if it does not perfectly reproduce the decaying behavior of the oscillations.
The split in OVS and non-OVS component however is reproduced perfectly.
Further, we provide missing plots for systems iv) and v) representing the pure learning-based scenario. 
In Fig. \ref{fig:lb} we demonstrate that the KKL-RNN is able to reproduce the dynamics of both systems and learns the correct split in OVS and non-OVS.
Fig. \ref{subfig:5_KKLRMM} further shows that the model can jointly learn the sine oscillations and VDP oscillator that is reconstructed from the sine oscillations. 
Fig. \ref{fig:lb_baselines} shows that the baselines struggle on these tasks.



\subsection{Runtimes}
All experiments are conducted on CPUs of an internal cluster. 
The runtimes for experiments i)-iii) are provided in Table \ref{t:hybrid}, the runtimes for experiments iv)-v) are provided in Table \ref{t:lb}.
\begin{table*}[h!]
	\centering 
	\caption{Mean of total runtimes in seconds for systems i)-iii) over 5 indep. runs.}
	\label{t:hybrid}
	\begin{tabular}{rccccc}
		\noalign{\smallskip} \hline \noalign{\smallskip}
		task & GRU & Residual Model  & Hybrid GRU & Filter & KKL-RNN (ours)  \\
		\hline
		i) & 283 & 276 & 264 & 262 & 535 \\
		ii)& 119 & 114 & 31  &  92 & 65  \\
		iii) & 3865   & 3967 & 3104 & 5310 & 8449  \\
		\noalign{\smallskip} \hline \noalign{\smallskip}
	\end{tabular}\quad
\end{table*}

\begin{table*}[ht]
	\centering 	
	\caption{Mean of total runtimes in seconds for systems iv)-v) over 5 indep. runs.}
	\label{t:lb}
	\begin{tabular}{rccccc}
		\noalign{\smallskip} \hline \noalign{\smallskip}
		task & GRU & Residual Model & Filter &Obs-GRU (ours) &KKL-RNN (ours)  \\
		\hline
		iv)  & 763  & 514 &- (-)&  556& 1078  \\
		v) & 1423  & 361  & 1436 &  118 & 207   \\
		\noalign{\smallskip} \hline \noalign{\smallskip}
	\end{tabular}\quad
\end{table*} 
\newpage

\FloatBarrier
\newpage
\section{Experimental details} \label{section:hyperparameters}
In this section, we will specify the configurations for each experiment.
This includes detailed descrpitions of the models as well as hyperparameters and training details. 

\subsection{Models} \label{section:mathematics}
In this section, we will add the necessary background on the models and data that we consider. 

\paragraph{Damped system: }
The data $x(t)$ are generated by adding a sine wave and a second sine oscillations that is exponentially damped over time. 
This yields the system
\begin{equation}
x(t)= \sin(0.1 t)+2\exp(-0.01 t)\sin(t).
\end{equation}
For the simulator, a simple sine wave with slight mismatch in the amplitude is chosen.
This yields
\begin{equation}
s(t)=0.7 \sin(0.1 t).
\end{equation}
The signals are obtained, by evaluating at $t=0, \dots 400$ with a discretization $\Delta t = 0.1$. Further the observation data are corrupted with white noise with variance $0.1$.  
\paragraph{Double torsion pendulum: }
For system ii), we consider the data from \citet{lisowski}.
We artificially add a transient component again $\Delta x$ via
\begin{equation}
\Delta x(t)=\exp(-0.5 t)\sin(50 t),
\end{equation}
evaluated at the interval $t = 0,1,2, \dots$. 
\paragraph{Van der Pol oscillator: }
For experiment iv), we consider a Van-der-Pol oscillator with additional control input. 
The differential equation for the Van-der-Pol oscillator with external force is given as
\begin{equation}
\begin{pmatrix}
\dot{x}(t) \\
\dot{y}(t) \\
\dot{u}(t) \\
\dot{v}(t)
\end{pmatrix}
=
\begin{pmatrix}
y \\
-x+a(1-x^2)y+bu \\
v \\
-\omega^2 u
\end{pmatrix},
\end{equation}
where $a,b$ and $\omega$ are system parameters. 
We choose $a=5, b=80, \omega=7.0$ and initial conditions 
\begin{equation}
\begin{pmatrix}
\dot{x}_0 \\
\dot{y}_0 \\
\dot{u}_0 \\
\dot{v}_0
\end{pmatrix}
=
\begin{pmatrix}
-2 \\
1 \\
0.31 \\
1
\end{pmatrix}.
\end{equation}
The system is evaluated at $t=5,\dots,100$ with step size $\Delta t=0.1$.

\subsection{Architecture details}\label{sec:arch}
Here, we will add details on the chosen architectures. 
This includes details on the networks and training details such as hyperparameter choice. 
\subsubsection{KKL model}\label{sec:KKL}
We will first describe the details of the KKL architecture.
This includes the options for the nonlinear transformation, the controllable pair and modeling variants for the non-OVS residuum. 
\paragraph{Nonlinear transformation: }
As proposed in \citet{9683277}, we consider an MLP with three layers and ReLU activation functions in between for the default scenario.
The number of neurons for each experiment will be documented in Sec \ref{sec:hyperparameters}. 
In the default scenario, we jointly propagate $z$ and $u$ through time and do not explicitly model $f_u$ but directly the observer. 
Due to the observer property, $u_{n+1}$ converges to $f_u(u_n)$. 
One advantage of this choice is that the whole rollout $z_0,\dots,z_N$ can be precomputed and the transformation can be applied in a batched manner making it efficient. 

However, sometimes direct access to $f_u$ might be required, e.g. in order to analyze fixed points or include symmetries. 
In this case, we provide the option to train $T^{\star}_{\theta}$ with an invertible neural network. 
The network is then built by stacking affine coupling layers \citep{DBLP:conf/iclr/DinhSB17}.
In particular, we stack 2 coupling layers, where the inner neural networks are modeled via 2 linear layers with ReLU activation functions. 
In contrast to the default setting, the transformation $T_{\theta}$ and the inverse $T_{\theta}^{\star}$ have to be computed in each time step, making it computationally more expensive. 
Furthermore, the invertible NN approach is less flexible since the number of neurons per layer has to match the latent dimensionality.  
Thus, if no direct access to the dynamis $f_u$ is required, the default setting is favorable. 
\paragraph{Controllable pair: }
The matrix $D_{\theta}$ is chosen as trainable diagonal matrix with sigmoid activation functions on the entries.
This ensures that $\lambda \in (0,1)$ for the eigenvalues $\lambda$.
However, of course this restricts the eigenvalues to positive ones.
Still, it worked well for our experiments.
Other activation functions such as tanh etc. can be easily incorporated.
It would also be possible to train the matrix freely as proposed in \citet{9683277}.
However, the pair $(D,F)$ would not meet the required properties in Thm \ref{Thm1} by design anymore.  
\subsubsection{Remaining components}
Here, we will describe the setting for the remaining components, in particular the non-OVS residuum.
\paragraph{Non-OVS residuum: } 
As stated in the method section, we model the non-OVS part via
\begin{equation}
v_{n+1}=f_{\theta}^v(u_n,v_n,y_n),
\end{equation}
where $f_{\theta}^v$ is modeled as GRU architecture. 
For efficiency reasons, we omit to provide $u_n$ as an input. 
Note that this is possible without loss of generality. 
However, the implementation of the method additionally contains the option to provide the simulator as a control input to $f_{\theta}^v$. 
\paragraph{Exponentially damped non-OVS residuum: }
In order to obtain a non-OVS component that vanishes over time, we propose to exponentially damp the observations. 
This is achieved by constructing a non-linear observation model $g$.
The latent states evolve according to the GRU transitions in Eq.~\eqref{eq:GRUequation}.
However, the linear observation function $g(x)=U_o x$ (cf. Eq.~\eqref{eq:obs}) is replaced by a non-linear observation model.
In particular, we consider 
\begin{equation}
g_{\textrm{damp}}(x)=a \exp(-\textrm{softplus}(b)t_n)\tanh(U_0 x),
\end{equation}
with trainable parameters $a$, $b$, a trainable linear model $U_0$ and the time interval $t=t_0, \dots, t_n$.

\subsection{GRU architecture}
We consider GRUs for our non-OVS residuum as well as for the baselines. 
We consider GRU dynamics $f$ as presented in Eq. \eqref{eq:GRUequation} with output feedback $y$ and control input $i$. 
The hidden GRU states are mapped to the observations via the observation model $g$ (cf. Eq. \eqref{eq:obs}).

\paragraph{Warmup: }
The hidden states of recurrent architectures such as GRUs are usually obtained with a short warmup phase of length $Rn$, where the architecture receives the observations as an input as described in Eq. \eqref{eq:warmup}.
The outputs are initialized with the observations on that horizon and therefore do not contribute to training. 
For the KKL observer, such a warmup phase is not required since it constantly receives the simulator outputs as an input.
However, the non-OVS residuum is initialized with standard GRU initialization via
\begin{equation}
y^v_{0:R}=\hat{y}_{0:R}-\left(g_{\theta}(u_n)\right)_{0:R}.
\end{equation}
This yields automatically that
\begin{equation}
y_{0:R}=\hat{y}_{0:R}. 
\end{equation}
Due to this architecture, the KKL observer contributes to training during the warmup phase but only via $s$, while the GRU does not contribute. 
However, this is balanzed by taking the MSE in the loss functions, which counterbalances different lengths of training trajectories. 
For the ablation study, all GRUs are initialized with exact data and start contributing to training after the warmup phase.

\paragraph{Hybrid GRU: }
For the hybrid GRU, we provide the simulator $s$ as control input $i$ to the GRU dynamics \eqref{eq:GRUequation}.

\paragraph{Residual model: }
For the residual model, we consider the sum of GRU predictions $r$ and simulations $s$.
In particular, we consider the transitions 
\begin{equation}
\begin{aligned}
x_{n+1}&=f(x_n,\tilde r_n) \\
\end{aligned}
\end{equation}
with corresponding observation model 
\begin{equation}
\begin{aligned}
r_n &=U_ox_n \\
y_n &=r_n+s_n.
\end{aligned}
\end{equation}
The warmup is obtained by feeding the residuum of data for $R$ steps.
This yields
\begin{equation}\label{eq:warmup}
\begin{aligned}
\tilde r_n = 
\begin{cases}
\hat y_n-s_n & \textrm{ for } n \leq R, \\
r_n=U_o x_n & \textrm{ for } n > R. 
\end{cases}
\end{aligned}
\end{equation}
The hyperparameters are optimized by minimizing $\Vert \hat y_n-y_n \Vert$.

\subsubsection{Filter design}\label{filter}
For the hybrid experiments, we reimplement a method similar to the hybrid model proposed in \citet{ensinger2023combining}.
It is implemented as follows:
\begin{itemize}
	\item Based on the properties of the simulator, the cutoff frequency is obtained, such that $L(s) \approx L(y)$.  
	\item The simulator signal is low-pass filtered with low-pass filter $L$, which yields $\tilde s = L(s)$.
	 This signal is stored.  
	\item During training and predictions the computation $\tilde s+H(y)$ is performed.
\end{itemize}
In contrast to their implementation, we apply forward-backward filtering provided by torch.
Further, the initialization slightly changes, since we filter the signals separately and add them.
Further, we precompute the low-pass filtered simulator signal and store it.  
The filter parameters are obtained with the help of \texttt{scipy.signal.iirfilter}. 
For the experiments here, we consider butterworth filters of order one. 
Thus, for a lowpass filter with cutoff frequency $\omega$, we call \texttt{iirfilter(N=1,Wn=$\omega$,rp=None,rs=None,btype="lowpass",analog=False,\\
ftype="butter",output="ba",fs=10)}.
Appropriate cutoff frequencies are obtained by analyzing the systems.
For training and predictions, we use the filters provided by \texttt{torchaudio}.
We use their implementation of forward-backward filtering \texttt{torchaudio.functional.filtfilt}, which first filters a signal and then filters it backwards.

\subsection{Hyperparameters}\label{sec:hyperparameters}
Here, we report the number of training steps, learning rates, latent dimensions, length of the warmup phase and additional parameters as cutoff frequency of the filters, damping factor etc. 
All models are trained with Adam optimizer. 
For all experiments, we train on subtrajectories of the full trajectory and consider batch sizes of length 50. 
To make a fair comparison, we choose an identical total amount of latent states for each experiment. 
\paragraph{i) Damped system: }
All models are trained with the learning rate $10^{-3}$. 
All models are trained for 300 steps. 
All models are trained on subtrajectories of length 100, the GRUs model receive 50 steps as a warmup phase. 
For our hybrid KKL-RNN, we consider a 32-dimensional latent space and an MLP with 100 neurons. 
The non-OVS residuum has a 32-dimensional latent space.
For the other models, we consider a GRU with 64 hidden states.
For the Filter, we consider the cutoff frequency 0.05.
\paragraph{ii): Double-torsion pendulum: }
All models are trained with the learning rate $10^{-3}$. 
The GRU is trained with an additional scheduler that multiplies the learning rate with 0.05 after 800 steps.  
All models are trained for 250 steps, the GRU is trained for 1000 steps. 
All models are trained on subtrajectories of length 50, the GRUs receive 20 steps for warmup. 
For our hybrid KKL-RNN, we consider a 64-dimensional latent space and an MLP with 100 neurons. 
The non-OVS residuum is regularized with weight 0.5. 
The non-OVS residuum has a 32-dimensional space. 
All other GRUs have 96-dimensional latent states. 
For the Filter, we consider the cutoff frequency 0.1.

\paragraph{iii) Drill-string system: }
All models are trained with the learning rate $10^{-3}$. 
All models are trained for 250 steps, the GRU is trained for 300 steps. 
All models are trained on subtrajectories of length 500, the GRUs receive 50 steps for the warmup phase.
For our hybrid KKL-RNN, we consider a 64-dimensional latent space and an MLP with 100 neurons. 
The non-OVS residuum is regularized with weight 0.5. 
The non-OVS residuum has a 32-dimensional space. 
For all other models, we consider GRUs with 96-dimensional hidden states. 
For the Filter, we consider the cutoff frequency 0.05.

\paragraph{iv) Van-der-Pol oscillator: } 
All models are trained with the learning rate $10^{-3}$.  
All models are trained for 500 steps, the GRU is trained for 800 steps. 
All models are trained on subtrajectories of length 200, the GRUs receive 20 steps for the warmup phase.
For our KKL-RNN, we consider a 32-dimensional latent space and an MLP with 100 neurons. 
The non-OVS part is regularized with weight 0.5. 
The non-observable GRU residuum has a 32-dimensional space. 
For all other models, we consider GRUs with 64-dimensional hidden states. 

\paragraph{v) Double torsion system: }
For the Filter baseline, we report the results provided in \citep{ensinger2023combining}.
For system v), we first pretrain the simulate substitute. 
To this, end the training signal is downsampled with a rate of 2 and low-pass filtered. 
On this signal, a GRU with 32 hidden dimensions is trained for 1000 training steps with a learning rate of $10^{-3}$. 
The trajectory is split into subtrajectories of length 125, 50 steps are used for warmup.
After training, the signal is upsampled with a rate of 2 again via the torch method \texttt{torch.nn.Upsample(scale\_factor = 2, mode="linear", align\_corners=False)}. 
The pretrained signal is then used as an input for the Residual model, the KKL-RNN and the Obs-GRU, while the GRU is trained in the standard setting on the whole input. 
All models are trained with a learning rate of $10^{-3}$.
KKL-RNN and Obs-GRU are trained on 150 steps, the Residual model is trained on 500 steps, the GRU is trained on 1151 steps. 
For all models, we consider subtrajectories of length 100 and for all GRUs, we consider 50 steps for the warmup phase. 
For our KKL-RNN, we consider a 32-dimensional latent space and an MLP with 100 neurons. 
The non-observable GRU residuum contains 32 hidden states. 
The residual model consists of a GRU with 64 hidden states. 
The Obs-GRU consists of a GRU with 64 latent states, the standard GRU has 96 latent states.

\bibliography{bib}

 